\documentclass[11pt]{article}

\usepackage{amsmath, amssymb}
\usepackage{graphicx}
\usepackage{caption}
\usepackage{subcaption}
\usepackage{multirow}
\usepackage[numbers]{natbib}
\usepackage{hyperref}
\usepackage{geometry}
 \usepackage{tikz}
 \usetikzlibrary{shapes.geometric, arrows.meta, positioning}

\title{\textbf{The Bulldozer Technique: Efficient Elimination of Local Minima Traps for APF-Based Robot Navigation}}

\author{
Mohammed Baziyad$^{1}$,
Manal Al Shohna$^{2}$,
Tamer Rabie$^{2}$ \\
\small $^{1}$Smart Automation and Communication Technologies (SACT),\\
\small Research Institute of Sciences and Engineering, University of Sharjah, UAE \\
\small $^{2}$Computer Engineering Department, University of Sharjah, UAE \\
\small \texttt{mbaziyad@sharjah.ac.ae, U19104878@sharjah.ac.ae, trabie@sharjah.ac.ae}
}

\date{} 

\begin{document}

\maketitle

\begin{abstract}
Path planning is a fundamental component in autonomous mobile robotics, enabling a robot to navigate from its current location to a desired goal while avoiding obstacles. Among the various techniques, Artificial Potential Field (APF) methods have gained popularity due to their simplicity, real-time responsiveness, and low computational requirements. However, a major limitation of conventional APF approaches is the local minima trap problem, where the robot becomes stuck in a position with no clear direction toward the goal. This paper proposes a novel path planning technique, termed the \textit{Bulldozer}, which addresses the local minima issue while preserving the inherent advantages of APF. The Bulldozer technique introduces a backfilling mechanism that systematically identifies and eliminates local minima regions by increasing their potential values, analogous to a bulldozer filling potholes in a road. Additionally, a ramp-based enhancement is incorporated to assist the robot in escaping trap areas when starting within a local minimum. The proposed technique is experimentally validated using a physical mobile robot across various maps with increasing complexity. Comparative analyses are conducted against standard APF, adaptive APF, and well-established planning algorithms such as A*, PRM, and RRT. Results demonstrate that the Bulldozer technique effectively resolves the local minima problem while achieving superior execution speed and competitive path quality. Furthermore, a kinematic tracking controller is employed to assess the smoothness and traceability of the planned paths, confirming their suitability for real-world execution.
\end{abstract}

\section{Introduction}\label{section-1-introduction}


Path planning is a fundamental problem in mobile robotics. It focuses on generating a feasible and efficient path that enables a robot to move from an initial position to a desired goal within a given environment. It constitutes a core component of a robot navigation system, which typically integrates localization, path planning, and control to enable autonomous motion. While localization estimates the robot’s pose and control ensures accurate trajectory execution, path planning bridges perception and actuation by providing a safe and goal-directed motion reference. Therefore, reliable path planning is essential for achieving robust and efficient autonomous navigation in real-world environments \cite{1,2}.

Path planning techniques are broadly categorized into two main classes: grid-based and sampling-based methods. Grid-based approaches discretize the environment into a fixed grid, where each cell is evaluated for occupancy and traversability. Classical algorithms such as A* and Dijkstra’s algorithm fall under this category and are known for their completeness and optimality guarantees on discrete maps. On the other hand, sampling-based techniques, such as the Probabilistic Roadmap (PRM) and Rapidly-exploring Random Tree (RRT), construct a graph or tree by randomly sampling points in the configuration space and connecting them based on feasibility. These methods are particularly advantageous in high-dimensional or continuous spaces where grid discretization becomes computationally prohibitive. Each class offers distinct trade-offs in terms of computation, memory usage, and path optimality, making the choice dependent on the specific application and environment characteristics \cite{4}.

The essential objective of any path planning algorithm is to generate a collision-free path from a start location to a goal while optimizing certain criteria—most commonly, minimizing the {path length} and ensuring {computational efficiency}. However, achieving both objectives simultaneously often involves a trade-off. For instance, {grid-based techniques} rely heavily on the resolution of the grid; using a finer resolution improves path precision and can result in shorter and smoother paths, but it also significantly increases computational time and memory requirements. Conversely, coarser grids reduce computation time but may yield suboptimal or even infeasible paths. Similarly, {sampling-based methods} such as PRM and RRT face a trade-off between execution speed and path quality. While fewer samples accelerate the planning process, they can result in longer, jagged, or incomplete paths. Increasing the sample density improves the likelihood of finding an optimal path but comes at the cost of longer processing times. Thus, a well-designed planner must carefully balance these objectives based on the complexity of the environment and the real-time constraints of the application.

Artificial Potential Field (APF) methods offer an elegant solution to the trade-off between execution speed and path optimality. Unlike grid-based or sampling-based techniques, APF does not require explicit map discretization or the construction of global graphs. Instead, it models the robot's environment using a continuous scalar potential field, where the goal exerts an attractive force and obstacles apply repulsive forces. The robot follows the negative gradient of this field, naturally moving toward regions of lower potential. This continuous formulation allows APF to produce smooth and goal-directed paths with minimal computational overhead, making it highly suitable for real-time applications. Its ability to react instantaneously to the potential field dynamics enables fast and responsive path generation, while the gradient descent behavior inherently encourages direct and efficient paths toward the goal.

Despite its simplicity and computational efficiency, the primary limitation of the APF method lies in the occurrence of {local minima traps}. These traps arise when the combined attractive and repulsive forces acting on the robot result in a net force of zero at a location that is not the goal. In such situations, the robot becomes trapped and is unable to progress toward its target, despite the absence of any physical collision with obstacles. Local minima are particularly common in environments with concave obstacles, narrow passages, or multiple closely spaced barriers. Since the robot follows the gradient of the potential field, it has no mechanism to escape such equilibrium points without external intervention or field modification. This limitation significantly affects the reliability of APF in complex environments and motivates the development of enhanced techniques to detect and resolve these undesired conditions.

To address the local minima trap problem while preserving the core advantages of the APF method, this paper introduces a new path planning approach that builds on the idea of modifying the potential landscape in a structured and adaptive manner. The rationale behind the proposed technique is inspired by a physical analogy: just as a bulldozer fills holes in a damaged road to ensure smooth travel, the method systematically identifies and eliminates low-potential regions that act as traps for the robot. Rather than relying on random perturbations or complex re-planning mechanisms, the approach treats local minima as repairable defects in the potential field, raising their values in a controlled way until a smooth descent path to the goal is restored. This strategy ensures that the robot continues to follow a natural gradient toward its destination without being hindered by artificial valleys or dead ends. The method aims to retain the simplicity, speed, and reactivity of APF, while introducing an intelligent mechanism for overcoming its primary shortcoming.

The proposed technique operates by first constructing a potential field using attractive and repulsive forces using a new suitable modeling strategy. The field is then segmented into non-overlapping blocks, and the average potential of each block is calculated. A block is considered a local minimum candidate if its average potential is lower than all eight of its neighboring blocks and it is not the goal block. Once identified, these local minima blocks are iteratively eliminated through a process called \textit{backfilling}, where their potential values are incrementally increased to remove the trap. This iterative process continues until no further local minima are detected. Additionally, when the robot starts inside a local minima region, a \textit{ramp mechanism} is applied. In this case, potential values are gradually assigned across the trapped area in a descending pattern that leads out of the trap, guiding the robot upward and away from the local minimum. The path is then generated by selecting the lowest potential neighboring block at each step, with Gray Wolf Optimization (GWO) used to identify the optimal point within each selected block, ensuring smooth and efficient progression toward the goal.

The main contributions of this paper can be summarized as follows:

\begin{enumerate}
    \item \textbf{A novel path planning technique – Bulldozer:}  
    A new APF-based path planning technique, termed \textit{Bulldozer}, is proposed to address the long-standing local minima trap problem in Artificial Potential Field methods, while preserving their computational efficiency and simplicity.

    \item \textbf{Precomputed APF modeling strategy:}  
    A new APF modeling strategy suitable for the proposed technique is introduced. Unlike conventional APF methods that rely on instantaneous gradient computation, the proposed approach precomputes the potential field over the entire map, with obstacles modeled using Gaussian functions. This formulation enables post-processing and field manipulation, which are not possible in standard gradient-based APF formulations.

    \item \textbf{Local minima detection and elimination strategy:}  
    A systematic block-based segmentation and backfilling mechanism is developed to detect and eliminate local minima areas in the potential field by increasing their potential values iteratively.

    \item \textbf{Ramp-based escape enhancement:}  
    An additional escape mechanism, called the \textit{ramp}, is introduced to help the robot escape traps when its initial position is inside a local minima. This technique ensures a gradual transition toward goal-directed motion.

\end{enumerate}

The rest of this paper is organized as follows: 
Section~\ref{section-3-literature-review} provides a comprehensive literature review, highlighting existing path planning techniques and identifying gaps that motivate this research.  
Section~\ref{section-5-the-proposed-design} introduces the proposed Bulldozer path planning technique, detailing its rationale, objective formulation, local minima handling strategy, and path planning process. Section~\ref{section-8-implementation-and-results} presents the experimental results, including comparisons with standard APF, adaptive methods, popular techniques such as A*, and PRM, as well as an analysis of path traceability and smoothness.   
Finally, Section~\ref{section-9-conclusions} concludes the paper with a summary of findings and closing remarks.

\section{Literature Review}\label{section-3-literature-review}

This section reviews existing research on mobile robot path planning, with particular emphasis on APF–based methods and metaheuristic optimization techniques. The literature encompasses a wide range of approaches developed to improve path feasibility, efficiency, and robustness under different environmental conditions and robotic platforms. To provide a structured overview, prior work is discussed according to methodological categories, including classical planning methods, APF enhancements, swarm-based optimization techniques, and advanced metaheuristic algorithms. This review highlights representative contributions within each category and establishes the context for the proposed approach.

\subsection{Improvements on APF}

The APF technique has been extensively improved to address local minima and unreachable goal problems. Several studies \cite{7,13,21} introduced regulative factors, escape forces, and adaptive step lengths to overcome these limitations. The augmented reality technique \cite{14} constructed virtual walls to completely close local minimum areas, preventing entrapment rather than requiring escape mechanisms. Temperature-based strategies \cite{22} implemented tempering and annealing processes to regulate attractive and repulsive forces dynamically. When local minima are detected, temperature increases sharply to amplify repulsive forces, enabling the robot to escape. Wall-following algorithms \cite{23} provide alternative solutions by switching behaviors when encountering obstacles, though this increases path length.

For Autonomous Underwater Vehicles (AUVs), the Regular Hexagon Guided technique \cite{15} addresses local minima by creating virtual hexagons near destinations, while distance correction factors solve the Goal Non-Reachable with Obstacle Nearby (GNRON) problem. Dynamic obstacle avoidance considers relative velocity components between robots and obstacles to determine optimal movement direction. Recent approaches \cite{36,66} integrate Deep Reinforcement Learning with APF, using traction forces and reward functions to optimize future paths and modify gravitational potential functions for smoother navigation.

Combining APF with other algorithms has proven effective for various robotic platforms. APFintegration with Ant Colony Optimization (ACO) \cite{18,19,51,65,73} leverages ACO's global search capability while APF provides local reactivity. Enhanced repulsive potential fields using exponential functions and adaptive pheromone update strategies address local minima and improve path quality. APF and Particle Swarm Optimization (PSO) combinations \cite{28} incorporate relative speed and acceleration between vehicles and obstacles, with virtual forces added when encountering local minima. APF-A* integration \cite{32,33,66} utilizes A* for global optimal paths while improved APF handles local planning, with optimized search processes minimizing unnecessary node evaluation.

For multi-robot systems, the Double Priority Joint Collision avoidance algorithm considers movement time and path length to prevent mutual collisions. The 4D Dynamic Priority RRT* combined with Heuristic APF \cite{58} introduces sub-target nodes as stepping stones to guide the robot to get out of stuck positions, ensuring coordinated flight timing in complex environments with static and moving threats.

\subsection{PSO Enhancements}

PSO has demonstrated superior performance in path length and quality across various experiments \cite{11}. Fractional-Order PSO \cite{27} introduces disturbance terms in velocity calculations, relating next iteration velocity to previous three iterations to improve convergence and enable thorough search space exploration. Improved localized PSO \cite{29} achieves better inertia weights, acceleration factors, and prevents local minima through Gaussian distribution for particle diversity. Chaos-based initialization, adaptive varying parameters, and mutation updates \cite{30} effectively improve convergence speed and solution optimality for robotic path planning under multiple constraints.

Hybrid PSO approaches combine PSO with Bald Eagle Search \cite{34}, Simulated Annealing \cite{39}, Coyote Optimization \cite{59}, and Cultural algorithms \cite{72}. These combinations balance global exploration and local exploitation through techniques such as chaotic initialization, adaptive parameter tuning, fitness-based switching, and probabilistic inertia weight updates inspired by the Metropolis rule. The RRT*-connect-PSO hybrid \cite{42} employs sampling-based methods for exploration and PSO for exploitation, with intermediary trees added in unexplored areas to improve path refinement.

\subsection{Advanced Metaheuristic Algorithms}

GWO technique has been enhanced through various strategies. Multi-strategy GWO \cite{71} incorporates Gaussian mutation mechanisms and spiral perturbation strategies with nonlinear Sigmoid convergence factors to maintain exploration-exploitation balance. Integration with Golden Sine Algorithm \cite{78} and lion optimizer dynamics \cite{48} improves convergence speed and precision. These modifications enable faster convergence while avoiding local minima through adaptive weights and relocation methods.

Whale Optimization Algorithm (WOA) improvements \cite{26,47} utilize erratic mapping for better population initialization and Corsi variation strategy for arbitrary perturbation updates. Enhanced Sparrow Search Algorithm \cite{62} employs fitness variation measures to detect stagnation, applying Gaussian random walk techniques to explore new possibilities. The Self-Adaptive Hybrid Bald Eagle Search \cite{60} dynamically adjusts search strategies using self-adaptive factors and Pigeon-Inspired Optimization concepts, with Bézier curves smoothing paths to reduce local minima risk.

Snake Optimization Algorithm (SOA) enhanced with Elite Opposition-Based Learning \cite{52} and improved Dung Beetle Optimizer \cite{75,80} incorporating cubic chaos mapping, adaptive t-distribution mechanisms, and spiral foraging strategies demonstrate superior performance in achieving faster convergence and enhanced robustness in dynamic, obstacle-rich environments. These algorithms address premature convergence through population diversity techniques and optimal dimension-wise mutation strategies.

\subsection{Other Optimization Techniques}

Cuckoo Search Algorithm adaptations \cite{25,41} replace fixed step sizes with adaptive ones, tuning both step size and probability factor parameters to balance exploration and exploitation stages. Enhanced Genetic Algorithms \cite{31,61} combined with Firefly algorithms employ selection, crossover, and mutation operations to escape local optima, while dynamic mutation and crossover operators in Genetic Algorithm (GA) integrated with Fast-Discrete Clothoid Curves generate safe paths for robots.

Improved Ant Colony Optimization with Q-Learning \cite{64} incorporates elite ant strategies, dynamic pheromone updates, and novel heuristics guided by third-order Bessel curves. Dual-phase planning approaches \cite{67} integrate blockage factors into ACO heuristic functions with conflict resolution mechanisms for multi-warehouse robots. Improved Artificial Bee Colony (ABC)\cite{68} leverages crowding distance strategies, Pareto optimal strategies, and external archive pruning to ensure solution diversity and prevent premature convergence.

Bacterial Foraging Optimization enhanced with Simulated Annealing \cite{74} incorporates acceptance of suboptimal solutions temporarily, enabling escape from local minima while adhering to COLREGs rules for robots. These diverse approaches demonstrate the continuous evolution of metaheuristic algorithms to address fundamental challenges in mobile robot path planning across various platforms and environments.

Despite the extensive efforts reported in the literature, most existing approaches mitigate the limitations of Artificial Potential Field methods by introducing external mechanisms, hybrid planners, or auxiliary optimization strategies. In contrast, the approach proposed in this paper operates directly on the potential field itself by modifying its structure in a controlled manner, rather than relying on additional planning or optimization frameworks. This direct field-level manipulation preserves the inherent advantages of APF methods, including computational efficiency, simplicity, and real-time applicability, while effectively addressing the local minima problem. As a result, the proposed technique maintains the core strengths of classical APF-based navigation while extending its robustness in challenging environments.

\section{The Proposed
Technique}\label{section-5-the-proposed-design}

\subsection{The Proposed
Rationale}\label{the-proposed-method}

APF techniques have long been utilized in robotic path planning due to their simplicity, computational efficiency, and effectiveness in real-time obstacle avoidance. The core idea behind APF is to model the robot's environment as a potential field, where obstacles exert repulsive forces, and the goal generates an attractive force. This method offers a direct and intuitive approach for guiding a robot from its starting position to the goal while avoiding collisions. The computational simplicity of APF allows it to be implemented even on resource-constrained robotic systems, making it ideal for a wide range of applications.

\begin{figure}[h]
    \centering
    \begin{subfigure}{0.48\textwidth}
        \centering
        \includegraphics[height=5cm, width = 5cm]{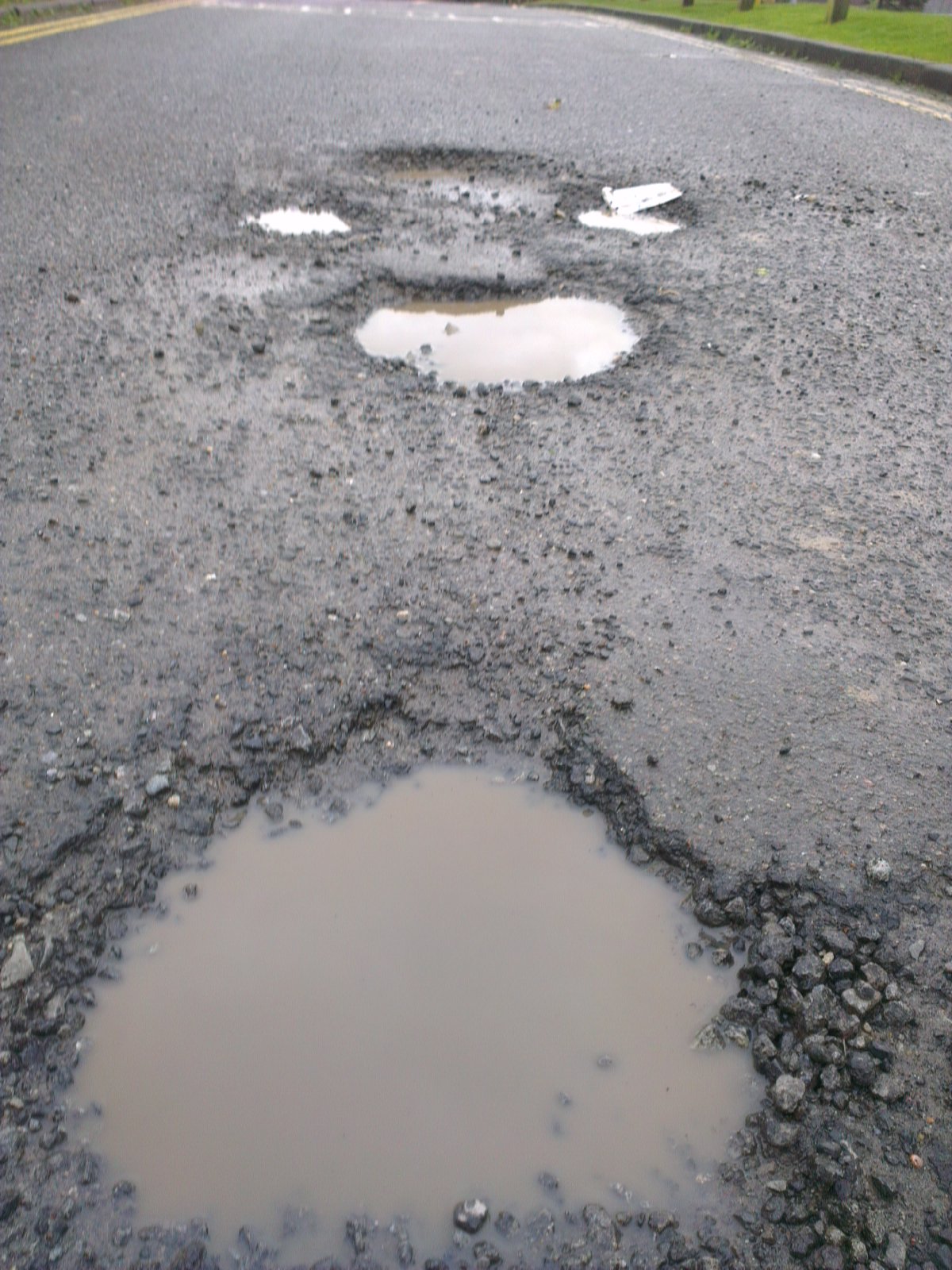}
        \caption{}
        \label{fig:local_minima}
    \end{subfigure}
    \begin{subfigure}{0.48\textwidth}
        \centering
        \includegraphics[height=5cm, width = 5cm]{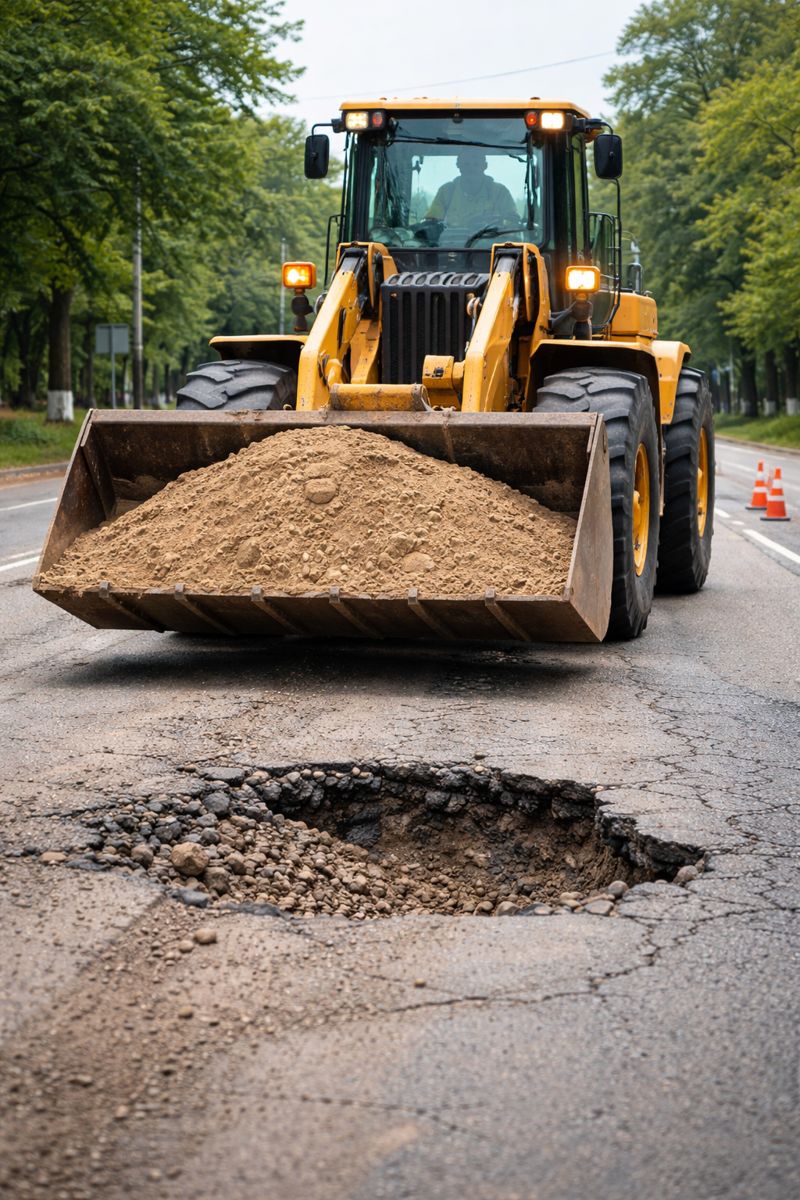}
        \caption{}
        \label{fig:bulldozer_elimination}
    \end{subfigure}
    \caption{Conceptual illustration of the proposed bulldozer technique for eliminating local minima in artificial potential field–based navigation: (a) presence of local minima modeled as pothole-like traps, and (b) flattening of the potential landscape through the bulldozer operation.}
    \label{fig:rationale}
\end{figure}

However, despite its advantages, APF techniques suffer from a critical weakness: the local minima trap problem. A local minimum occurs when the robot finds itself in a position where the attractive force toward the goal is balanced by the repulsive forces from surrounding obstacles, resulting in a zero net force. When this happens, the robot becomes stuck, unable to progress towards its goal. This issue is particularly problematic in environments with concave obstacles or narrow passages, where local minima are more likely to occur. The local minima trap problem significantly limits the robustness of traditional APF techniques, motivating the need for enhanced solutions.

The proposed "Bulldozer" technique draws inspiration from the analogy of a bulldozer backfilling potholes on a highway to create a smooth and navigable surface as shown in Figure \ref{fig:rationale}. In the context of robotic path planning, local minima in the potential field represent these problematic potholes where a robot might get stuck. The Bulldozer technique addresses this issue by systematically identifying local minima areas within the potential field and dynamically modifying their potential values. By effectively 'filling' these low-potential regions, the method transforms local minima into passable areas, ensuring uninterrupted progress toward the goal while preserving the inherent simplicity and efficiency of APF methods.

\subsection{The Proposed System Phases}
\begin{figure}[h]
\centering
\begin{tikzpicture}[
    node distance=9mm and 15mm,
    font=\small,
    startstop/.style={ellipse, draw, align=center, minimum width=26mm, minimum height=8mm},
    block/.style={rectangle, draw, rounded corners, align=center, minimum width=52mm, minimum height=8mm},
    decision/.style={diamond, draw, aspect=2, align=center, inner sep=1pt, minimum width=44mm, minimum height=10mm},
    subblock/.style={rectangle, draw, rounded corners, align=left, minimum width=58mm, minimum height=18mm},
    line/.style={-Latex, thick}
]

\node[startstop] (start) {Start};

\node[block, below=of start] (pf) {Compute potential field $V(\mathbf{x})$};
\node[block, below=of pf] (seg) {Segment $V(\mathbf{x})$ into\\ non-overlapping blocks};
\node[block, below=of seg] (search) {Search over blocks for local-minimum candidates};

\node[decision, below=of search] (min) {Local-minimum\\ block found?};

\node[subblock, right=of min, xshift=12mm] (elim) {%
\textbf{Ramp-based elimination}\\
Initialize $k \leftarrow 1$ (if first time)\\
Set boost $\Delta V_k$ (highest at $k{=}1$)\\
Increase potential in block: $V \leftarrow V + \Delta V_k$\\
Update ramp: $k \leftarrow k+1$, $\Delta V_{k+1}<\Delta V_k$
};

\node[block, below=of min] (path) {Extract path on the modified field};

\node[startstop, below=of path] (end) {End};

\draw[line] (start) -- (pf);
\draw[line] (pf) -- (seg);
\draw[line] (seg) -- (search);
\draw[line] (search) -- (min);

\draw[line] (min) -- node[above]{Yes} (elim);
\draw[line] (min) -- node[left]{No} (path);
\draw[line] (path) -- (end);

\draw[line] (elim.east) -- ++(12mm,0) |- (search.east);

\end{tikzpicture}
\caption{Flowchart of the proposed block-based local minima elimination strategy with a ramp mechanism. When a local-minimum block is detected, its potential is increased by a boost $\Delta V_k$; the first detected block receives the largest boost, and subsequent boosts decrease monotonically ($\Delta V_{k+1}<\Delta V_k$), creating a ramp-like potential rise that helps the robot escape minima traps.}
\label{fig:ramp_flowchart}
\end{figure}
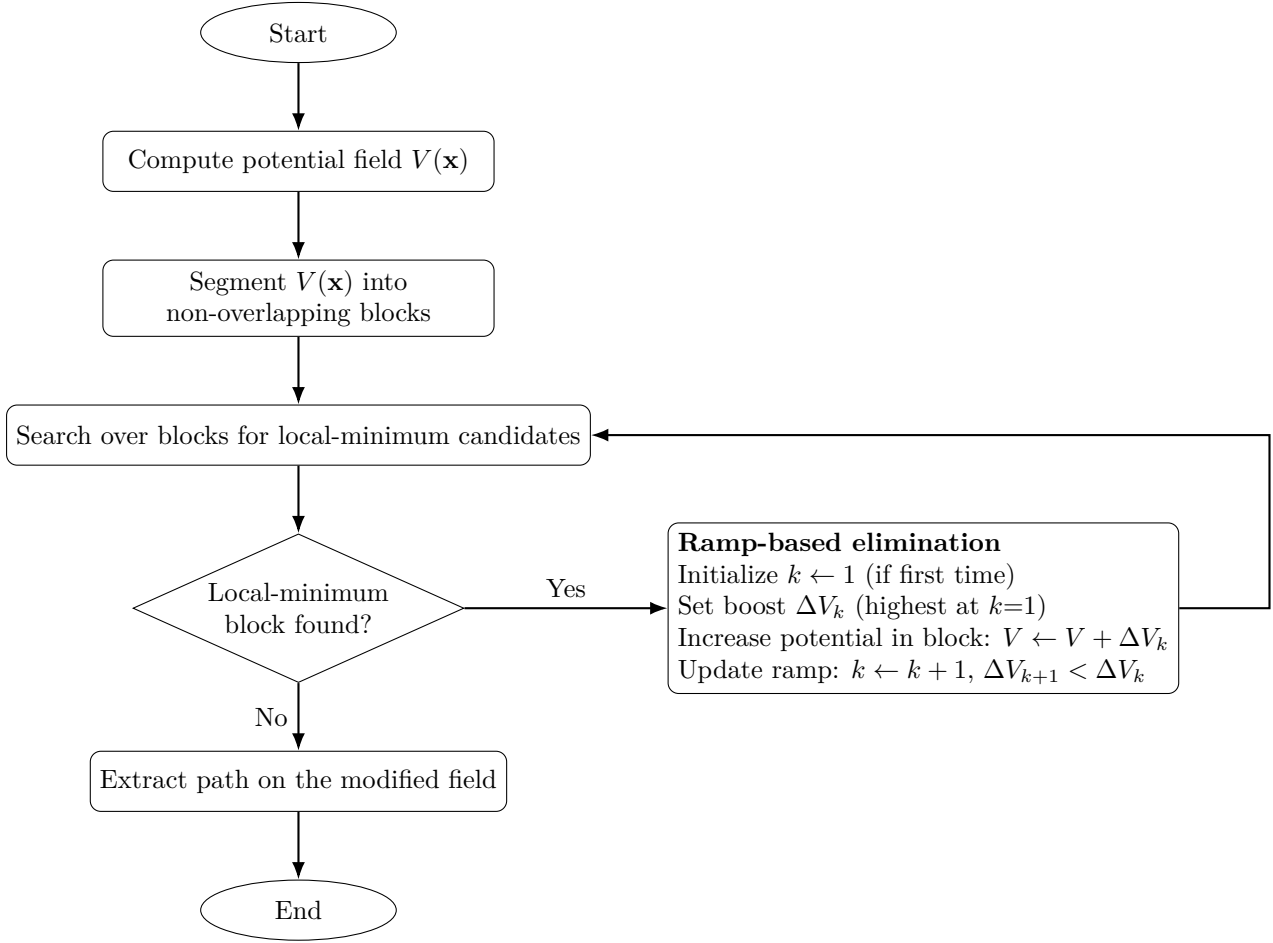

Figure~\ref{fig:ramp_flowchart} presents the core stages of the proposed Bulldozer path planning framework. Each stage is designed to address specific challenges in traditional APF methods, particularly the local minima problem. The phases are described in detail as follows:

\begin{enumerate}
    \item \textbf{Artificial Potential Field Construction:} In this phase, a classical potential field is generated over the map using both attractive and repulsive components. The goal point creates an attractive well, pulling the robot toward it, while each obstacle introduces a repulsive peak to prevent collisions. This field provides a preliminary navigation landscape where the robot can theoretically follow the gradient to reach its destination. However, in practice, this field may contain local minima traps where the robot gets stuck.

    \item \textbf{Identification of Local Minima Areas:} Once the potential field is constructed, the next step involves scanning the field to identify potential trap regions. The map is segmented into non-overlapping blocks, and for each block, its average potential value is compared with its neighboring blocks. If a block has the lowest value compared to all eight surrounding blocks, and it is not the goal, it is considered a candidate local minimum. This phase is essential to proactively detect problematic areas before they interfere with the robot's path.

    \item \textbf{Elimination of Local Minima:} After detecting the local minima regions, a backfilling process is applied to systematically eliminate these traps. This involves increasing the potential value of the identified blocks to a level higher than their neighbors, essentially flattening or raising the trap zone. The process is iterative—after each pass, new neighboring blocks may become local minima and are similarly treated. This continues until no new minima are found. This mechanism mimics a bulldozer backfilling potholes in a road, hence the name of the technique.

    \item \textbf{Path Planning:} With a clean, trap-free potential field, the robot can now plan its path effectively. The path planning algorithm selects the next step by moving toward the neighboring block with the lowest potential value, progressing toward the goal. If a direct connection is blocked by an obstacle, a local search such as Dijkstra’s algorithm is applied to detour around the obstruction. This final phase ensures that the robot reaches its target using a smooth, safe, and computationally efficient route.
\end{enumerate}




\subsection{Formulating the Objective Function}\label{formulating-the-objective-function}

The main objective of path planning method is to find a free-collision
path from the start point to the destination. The proposed method
simulates the main concept of artificial potential field, which means
attractive and repulsive forces are presented in the environment. As the
robot should avoid obstacles and keep moving towards the target,
repulsive forces are produced by the obstacles and attractive forces are
emitted by the target point. In the proposed system, the attractive force is modeled using Euclidean distance. In other words, the attractive force is a function of the distance from the goal point. Therefore, by optimizing this function (finding the minium) iteratively over time, the robot should be able to move to the goal point. 

However, with the presence of obstacles in the environment, there is a need to add a term to the Euclidean term to act as a repulsive force which should make the robot avoid colliding into obstacles. Mathematically, this repulsive term should increase the potential field values around obstacles. Therefore, by optimizing (minimizing) this new function, the robot will keep moving towards the goal point, while avoiding approaching obstacles. In this proposed work, the Gaussian function is used to simulate the repulsive force. The reason is that the Gaussian function will give strong force (very high potential values) around the obstacles, while approximately zero influence within free-space areas. This design of APF is specific to the proposed system and is different from the standard APF technique. This modified version is believed to be simpler and therefore faster execution speed is expected.

\begin{figure}[h!]
    \centering
    \includegraphics[width=0.85\textwidth]{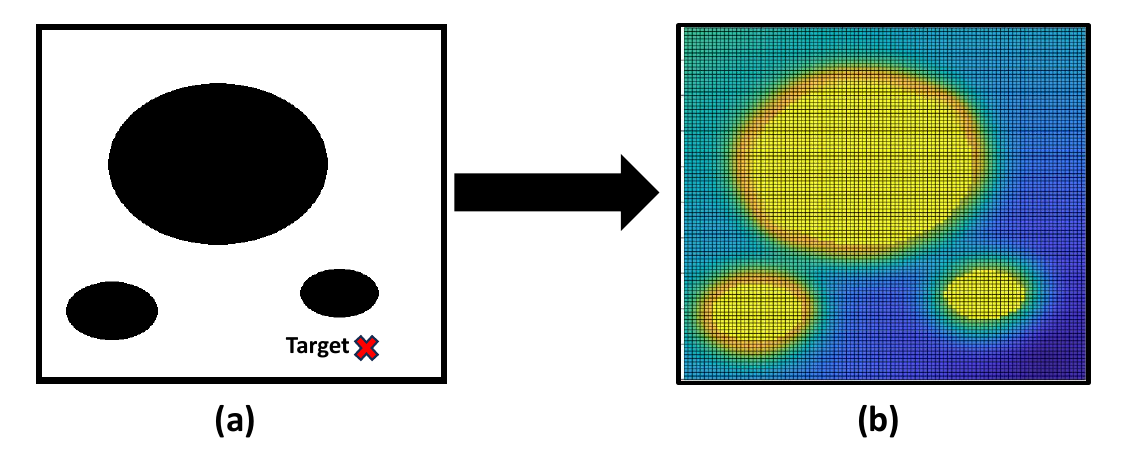} 
    \caption{Illustration of potential field generation: (a) the environment map with circular obstacles and a target point, and (b) the resulting potential field where high values are centered around obstacles and low values surround the goal.}
    \label{fig:potential_field_generation}
\end{figure}

The process of generating the potential field is visualized in Figure~\ref{fig:potential_field_generation}. Subfigure (a) shows a basic map containing three circular obstacles and a target location marked by a red “X”. This spatial setup is used to construct the potential field, as shown in subfigure (b), which maps the environment using the APF technique. In this field, regions surrounding obstacles are assigned high potential values (depicted in yellow), acting as repulsive zones to prevent collisions. Conversely, the goal point produces a low potential region (shown in dark blue), creating an attractive force that pulls the robot toward the target. The smooth transition from high to low potential across the field creates a navigable gradient that guides the robot in real time. This field structure is fundamental to the APF approach, forming the basis for subsequent stages of the proposed Bulldozer path planning technique.

\subsubsection{Mathematical Modeling}
The objective value $F$ at a given point in the map $(x_i,y_i)$ can be found as follows in general:

\begin{equation}
  F(x_i,y_i)\  = \ a.\ E(x_i,y_i)\  + \ b.\ \sum _{o=1}^K G(x_i,y_i)  
\end{equation}
\(E(x_i,y_i)\) is the Euclidean distance from the point \((x_i,y_i)\) to the
goal point \(p_{g}\). \(G(x_i,y_i)\) is the sum of the Gaussian values from all $K$ obstacles. The \(a\) and \(b\) factors are normalization functions.
\\
The Euclidean function $E(x_i,y_i)$ can be found as: 

\begin{equation}
\label{eq:at}
    E(x_i,y_i) = \sqrt{(x_{g} - x_{i})^{2} + (y_{g} - y_{i})^{2}}
\end{equation}
where $(x_{g},y_{g})$ are the coordinates of the goal point. As the robot is attracted by the goal point and keeps moving towards it,
the distance between the robot and the target point needs to be
minimized. When the robot arrives at the goal point this distance will
be zero, i.e., \(E\  = \ 0\).

On the other hand, $G(x_i,y_i)$ is a Gaussian function that represents the repulsive force emitted by an obstacle. The
following formula represents the repulsive force at $(x_i,y_i)$ emitted by an obstacle at $(x_o,y_o)$:
\begin{equation}
\label{eq:g}
G(x_i,y_i) = e^{\frac{(x_i - x_{o})^{2}}{2\sigma_{x}^{2}} + \frac{(y_i - y_{o})^{2}}{2\sigma_{y}^{2}}}
\end{equation}
where\(\ \sigma\ \)is the standard deviation. A Gaussian function has a
peak value that is needed to represent the high repulsive value when the
robot is close to the obstacle. The value of the function
decreases and almost reaches to zero for x values that are far away from
the obstacle, which can represent the lower repulsive force that affects
the robot while moving far away from the obstacle. In other words, the repulsive force increases when the robot is closer to the obstacle,
and gradually decreases as the robot moves far away from it.

As there are multiple obstacles in the environment, the objective
function has the summation of repulsive forces emitted from all
obstacles.

\begin{equation}
\label{eq:re}
G = \sum_{o = 1}^{k} G_{o}; \quad 1 < o < k
\end{equation}
$k$ represents the number of obstacles.
\\
Combining Eq. \ref{eq:at}, \ref{eq:g}, and \ref{eq:re}, the final APF function is achieved:

\begin{equation}
\label{eq:final}
  APF(x,y) =  \sqrt{(x_{g} - x_{i})^{2} + (y_{g} - y_{i})^{2}} + \sum_{o = 1}^{k}  e^{\frac{(x - x_{o})^{2}}{2\sigma_{x}^{2}} + \frac{(y - y_{o})^{2}}{2\sigma_{y}^{2}}} 
\end{equation}

\subsection{Identifying Local Minima Areas}\label{identifying-local-minima-areas}

One of the most critical steps in the proposed system is the identification of local minima areas in the field where the robot may become trapped due to a balance of repulsive and attractive forces. This subsection introduces the strategies used to detect such problematic regions. Two different approaches are explored: an analytical method that seeks to locate local minima using mathematical expressions and gradient analysis, and a practical heuristic approach, which is the adopted approach in this paper. The heuristic approach works by dividing the potential field into discrete blocks and evaluates potential values relative to neighboring regions. The objective of this phase is to provide a reliable mechanism to pinpoint and label local minima areas, forming the foundation for their subsequent elimination and enabling uninterrupted robot navigation toward the goal.

\subsubsection{Analytical Discussion}\label{analytical-method}

This approach attempts minimizing the APF equation presented in Eq. \ref{eq:final} analytically. While the APF equation is a smooth and continuous function, solving for the global minimum analytically is challenging due to the nonlinearity and multi-modal nature of the function. To analytically find the optimal path, we must solve:
\begin{equation}
    \nabla APF(x, y) = 0
\end{equation}
Computing the gradient of Equation~\ref{eq:final} gives:
\begin{equation}
\frac{\partial APF}{\partial x} = \frac{x - x_g}{\sqrt{(x - x_g)^2 + (y - y_g)^2}} + \sum_{o=1}^k \frac{x - x_o}{\sigma_x^2} \cdot e^{\left(\frac{(x - x_o)^2}{2\sigma_x^2} + \frac{(y - y_o)^2}{2\sigma_y^2}\right)}
\end{equation}

\begin{equation}
\frac{\partial APF}{\partial y} = \frac{y - y_g}{\sqrt{(x - x_g)^2 + (y - y_g)^2}} + \sum_{o=1}^k \frac{y - y_o}{\sigma_y^2} \cdot e^{\left(\frac{(x - x_o)^2}{2\sigma_x^2} + \frac{(y - y_o)^2}{2\sigma_y^2}\right)}
\end{equation}
Solving this system for $(x, y)$ is non-trivial, especially when multiple obstacles are present, as the exponential terms dominate locally and can create complex surface shapes.
\\
\\
{\bf{Simple Case: One Circular Obstacle}}
\\
Let us consider a simplified example: a single circular obstacle centered at the origin $(0, 0)$ and a goal at $(2, 0)$. Assume:
\\
- $\sigma_x = \sigma_y = \sigma$\\
- The APF becomes:\\
\begin{equation}
    APF(x, y) = \sqrt{(x - 2)^2 + y^2} + e^{\left(\frac{x^2 + y^2}{2\sigma^2}\right)}
\end{equation}
We compute the gradient:
\begin{align}
\frac{\partial APF}{\partial x} &= \frac{x - 2}{\sqrt{(x - 2)^2 + y^2}} + \frac{x}{\sigma^2} e^{\left(\frac{x^2 + y^2}{2\sigma^2}\right)} \\
\frac{\partial APF}{\partial y} &= \frac{y}{\sqrt{(x - 2)^2 + y^2}} + \frac{y}{\sigma^2} e^{\left(\frac{x^2 + y^2}{2\sigma^2}\right)}
\end{align}
To find the minima analytically, we solve:
\begin{equation}
\frac{\partial APF}{\partial x} = 0 \quad \text{and} \quad \frac{\partial APF}{\partial y} = 0
\end{equation}
For symmetry, try $y = 0$, which simplifies the system:
\begin{equation}
\frac{x - 2}{|x - 2|} + \frac{x}{\sigma^2} e^{\left(\frac{x^2}{2\sigma^2}\right)} = 0
\end{equation}
Even this reduced equation is transcendental (involves both algebraic and exponential terms), meaning it has {\bf no closed-form solution} and must be solved numerically.

This concludes that while the APF function is analytically differentiable, its minimization is rarely solvable in closed form due to the nonlinear interaction of attraction and repulsion components. Even in simplified settings, the resulting equations are transcendental and require numerical methods or meta-heuristic strategies to approximate the global minimum. This complexity reinforces the motivation behind techniques such as the proposed \textit{Bulldozer}, which handle local minima directly via potential surface shaping, rather than exact symbolic optimization.

\subsubsection{The Proposed Heuristic Method}

The heuristic method proposed in this work offers a simple yet effective approach to detect local minima regions in the potential field without requiring complex gradient analysis. Instead of relying on analytical computation of the field derivatives, the method segments the potential field map into a grid of non-overlapping square blocks. Each block represents a small portion of the field, and the average potential value within each block is computed.

To determine whether a given block is a local minimum, its average potential value is compared with those of its eight surrounding neighboring blocks (top, bottom, left, right, and the four diagonals). If the block under evaluation has a strictly lower average value than all its neighbors and is not the goal block itself, it is classified as a local minimum candidate.

This method is inspired by local topology checks commonly used in image processing and terrain analysis, where a central pixel or region is evaluated relative to its surroundings. The advantage of this approach lies in its simplicity and computational efficiency, making it suitable for real-time implementation on embedded systems. Moreover, the use of block-level comparison rather than pixel-level values adds robustness against noise and small fluctuations in the potential field.

Once all local minima blocks are identified, the proposed system proceeds to eliminate them through the backfilling process described in the next phase. This heuristic method ensures that the potential field is free of undesired traps that could otherwise prevent the robot from reaching its destination.

In this method the larger the block size the more accurate the solution
is, and the longer the computation time is. On the other hand, the
smaller the block size the lass accurate solution is, and the shorter
computation time is. There has to be a trade-off between the speed and
the quality when the map resolution is selected. 

\begin{figure}
    \centering
    \includegraphics[width=0.75\linewidth]{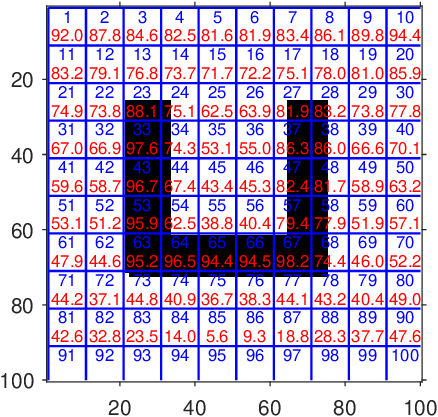}
    \caption{Block-wise analysis of the potential field for the U-shaped obstacle map. The figure displays a grid-based segmentation where each block is labeled with its block number (in blue) and its corresponding average potential value (in red). The average values are computed over all pixels in each block. The goal point is within block number 85. Blocks with very high average potentials typically correspond to obstacle regions, whereas blocks with lower values indicate more favorable areas for navigation. This representation supports the heuristic method for local minima detection by enabling comparison of a block's potential with its neighbors. The U-shape creates an enclosed region where a local minimum trap is likely to form, highlighting the need for a robust identification strategy. It is clear that block number 55 is a local minima since it has lower values than its surrounding 8 blocks.}
    \label{fig:u_obstacle_avg_blocks}
\end{figure}

Figure~\ref{fig:u_obstacle_avg_blocks} illustrates the identification of local minima regions step, where the potential field is segmented into non-overlapping square blocks. Each block is assigned a unique identifier (shown in blue) and its corresponding average potential value (shown in red below). This block-based representation enables a simplified and efficient heuristic for local minima detection.

A block is considered a local minima candidate if its average potential value is lower than that of all its surrounding neighbors expect for the block which includes the goal point, which is block number 85 in Figure \ref{fig:u_obstacle_avg_blocks}. For example, in Figure~\ref{fig:u_obstacle_avg_blocks}, block 55 has an average potential of 38.8, which is significantly lower than its eight surrounding neighbors—namely blocks 44 (67.4), 45 (43.4), 46 (45.3), 54 (62.5), 56 (40.4), 64 (96.5), 65 (94.4), and 66 (94.5). Since block 55 has the lowest average among its neighbors, it is flagged as a local minima candidate. This condition suggests that if a robot were navigating using the potential field, it might become trapped in this block unless the local minimum is addressed by the proposed Bulldozer method.

\subsection{Eliminate Local Minima Points}\label{eliminate-local-minima-points}

In this study we deal with local minimum points as holes. After these
holes are located, they have
to be backfilled. This operation is implemented manually by assigning
high values to local minima points enough to enable the robot to avoid
falling in them. The difference between the proposed method and other
path planning methods is the robot never gets into the local minimum
because the low values are replaced with high values before the planning
phase. Other planning methods in the literature the robot tries to jump out of local solution after being trapped in it.

The process of eliminating local minima in the proposed technique is inherently iterative. In each iteration, a candidate local minima block is identified based on its average potential value relative to its neighbors. Once a block is confirmed as a local minimum, its potential value is artificially increased—a process we refer to as backfilling—so that it no longer acts as a trap for the robot. The algorithm then re-evaluates the entire field to locate new local minima that may emerge as a result of the previous update.

Referring back to Figure~\ref{fig:u_obstacle_avg_blocks}, suppose block 55 is identified as a local minimum in the first iteration due to its relatively low average potential value. Once block 55 is backfilled and its potential is increased, block 56—previously not considered a minimum—may now have the lowest value among its new set of neighbors, including the updated block 55. In the subsequent iteration, block 56 is then treated as the new local minimum and also backfilled. This chain continues until all potential traps in the region are raised sufficiently and no local minimum remains. Such an iterative procedure ensures that not just individual low-potential zones, but entire trap regions are effectively eliminated from the potential field, thereby enabling uninterrupted navigation toward the goal.

\subsubsection{Ramp Mechanism for Trap Escape}

While the Bulldozer technique effectively eliminates local minima regions during the potential field adjustment phase, it is still possible for the robot to start within a local minima zone—especially in scenarios where the trap is part of the initial configuration. In such cases, the robot may still struggle to initiate movement toward the goal, as it starts from a locally flat or high-potential area with no clear descent direction.

To address this challenge, a complementary strategy referred to as the \textit{Ramp Mechanism} is proposed. The concept is inspired by the analogy of constructing a ramp that helps the robot ascend out of a depression before proceeding downhill toward the goal. The Ramp Mechanism works by creating a gradual transition in potential values within the local minima zone itself.

The implementation involves assigning a descending gradient of potential values starting from the innermost trap region outward toward the nearest non-trap blocks. In this way, the robot is encouraged to follow this artificially constructed slope, effectively guiding it out of the local minima. The first block identified as a local minimum is assigned a relatively higher potential value, and subsequent surrounding blocks are assigned progressively lower values to create a sloped ramp structure.

This gradual decline in potential ensures that even if the robot starts in a trap, it can follow a smooth gradient outward, avoiding stagnation and enabling it to join the main potential descent path toward the goal. The ramp concept, therefore, enhances the robustness of the proposed technique by offering a recovery mechanism from adverse starting conditions without affecting the simplicity or speed of the overall planning process.

\begin{figure}
    \centering
    \includegraphics[width=0.75\linewidth]{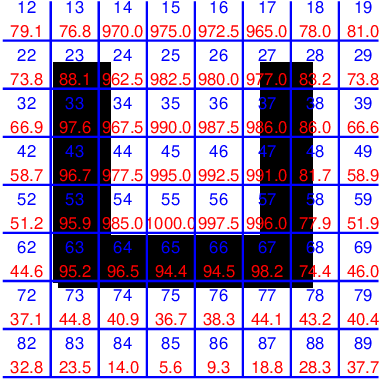}
\caption{A zoomed section showing the potential field distribution after applying the proposed Bulldozer technique with the ramp concept on the blocks shown in Figure \ref{fig:u_obstacle_avg_blocks}. Block 55, initially identified as a local minimum, is assigned a high potential value (1000), and surrounding blocks inside the U-shaped obstacle are gradually filled with decreasing values. This gradient forms a ramp-like structure that guides the robot out of the trap area while ensuring no new local minima are introduced. The robot can then easily navigate to the goal point located at block number 85.}
    \label{fig:ramp_result}
\end{figure}

Figure~\ref{fig:ramp_result} illustrates the state of the potential field after applying the proposed Bulldozer technique to eliminate local minima within the U-shaped obstacle region. In this implementation, the ramp concept is utilized to ensure that the robot can smoothly escape from any potential trap. As shown in the figure, block 55, which was initially identified as the core of the local minima trap, has been assigned a high potential value of 1000. This increase effectively transforms it into a repelling region that prevents the robot from getting stuck.

More importantly, the surrounding blocks within the U-shaped enclosure have also been modified. Their potential values are no longer low; instead, they gradually decrease as the robot moves away from block 55 and toward the exit. This smooth gradient forms a virtual ramp—high potential at the trap center and gradually lower values outward—guiding the robot out of the local minima region.

For instance, block 56 is now at 997.5, followed by 57 at 996.0, and so on. This cascade of decreasing values ensures that none of these blocks become local minima themselves. The entire U-shaped region has thus been neutralized as a trap zone through a continuous process of iterative filling and evaluation, demonstrating the effectiveness of the proposed method in creating a clean navigable path while preserving the simplicity and responsiveness of the original APF technique.

\subsection{Planning the Path}

After eliminating the local minima areas using the Bulldozer mechanism, the robot proceeds with the actual path planning phase. The proposed planner operates by iteratively selecting the nearest neighboring block that has the lowest potential value in the direction of the goal. This strategy enables the robot to follow a downhill gradient toward the goal without falling into previously identified traps. The block-based selection process ensures a smooth and progressive movement through the potential field.

\begin{figure}[htbp]
    \centering
    \includegraphics[width=0.5\textwidth]{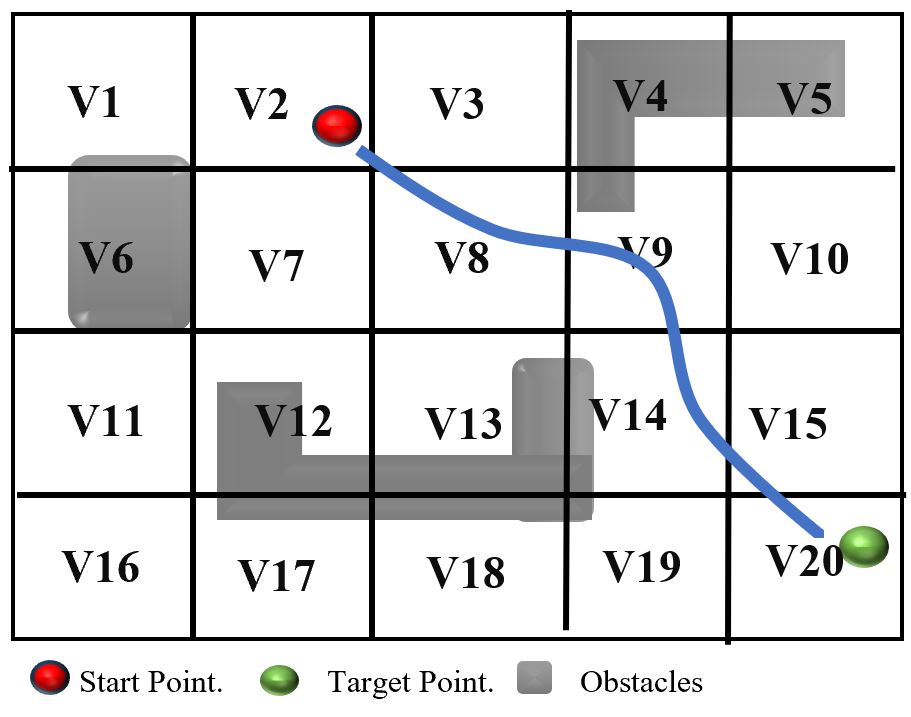}
    \caption{Planned path after eliminating local minima using blocks method. In each block, the GWO algorithm is adopted to find the optimum value and selects it as a path point.}
    \label{fig:planned_path_blocks}
\end{figure}

As illustrated in Figure~\ref{fig:planned_path_blocks}, the robot successfully navigates from the start position to the target point by passing through adjacent low-potential blocks while avoiding obstacles. The path is naturally shaped by the updated potential field and adheres to the structure of the map, ensuring efficient goal-directed motion.

\begin{figure}
    \centering
    \includegraphics[width=0.75\linewidth]{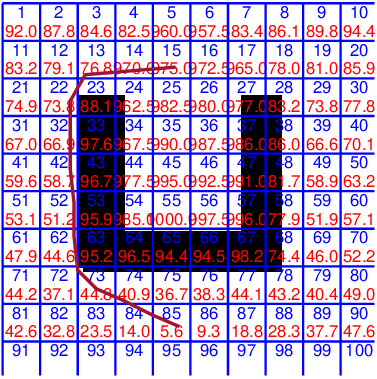}
\caption{Path generation using block-based potential field minimization. At each step, the next move is determined by selecting the neighboring block with the lowest average potential. Within that block, the GWO algorithm is employed to find the minimum potential point, which is selected as the next path node. The red curve shows the resulting planned path.}
    \label{fig:path_planning_gwo}
\end{figure}

Figure~\ref{fig:path_planning_gwo} illustrates the path planning phase after the local minima have been eliminated. In this approach, the robot starts at a given position and evaluates the surrounding blocks to determine the next movement direction. At each step, the algorithm selects the block with the lowest average potential value as the next destination.

To precisely locate the point to move within the selected block, the GWO algorithm is applied. The GWO algorithm searches within the candidate block to find the specific coordinate that corresponds to the minimum potential value. This approach provides fine-grained control, allowing the robot to avoid high-potential zones while navigating through low-cost valleys in the potential field.

As shown in Figure \ref{fig:path_planning_gwo}, the red curve represents the planned path connecting the GWO-selected optimal points across blocks. The path successfully follows a smooth and feasible trajectory from the starting point toward the goal, avoiding previously problematic trap zones.

\subsection{Dijkstra Algorithm for Local Detours}

In certain situations, the direct connection from the current block to the next lower-potential block may intersect with an obstacle. In such cases, the proposed technique employs a lightweight detouring mechanism using the Dijkstra algorithm to compute a local shortest path around the obstruction. Since this operation is only invoked for a small segment of the overall map, the computational overhead remains minimal and does not compromise the overall speed of the method.

By integrating Dijkstra’s algorithm selectively and only when necessary, the proposed approach ensures both obstacle avoidance and path continuity without deviating from its core design principle of efficiency. This hybrid mechanism enables the robot to maintain a balance between optimality and execution speed, especially in cluttered environments where direct transitions are occasionally blocked.

\section{Experimental
Results}\label{section-8-implementation-and-results}
This section examines the performance of the proposed technique. The primary objective is to assess the technique’s capability to successfully navigate toward the goal point in environments containing local minima traps. The performance is first compared against the standard APF method, which is known to struggle in such scenarios. Subsequently, the Bulldozer technique is benchmarked against other competitive approaches specifically designed to overcome local minima, with the comparison focusing on execution speed and path length. To further evaluate its robustness, the proposed method is also compared with general, well-established path planning algorithms that are typically immune to local minima issues. Moreover, since the technique is based on GWO, a comparison with other meta-heuristic is also performed. Finally, the smoothness and tracking performance of the generated paths are analyzed using a basic kinematic controller.

\begin{figure}[htbp]
    \centering
    \includegraphics[width=0.7\linewidth]{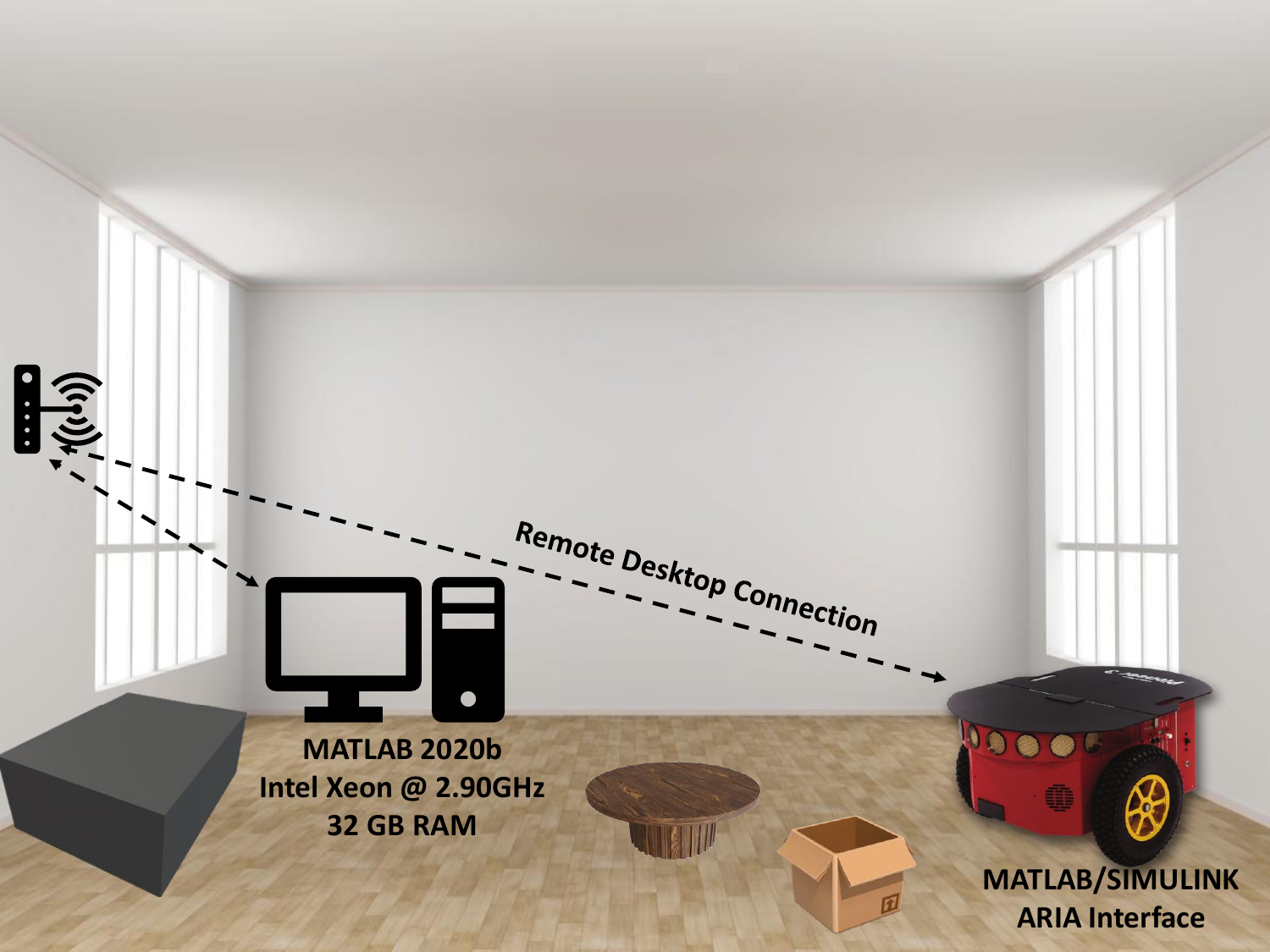}
    \caption{Experimental setup}
    \label{fig:experimental_setup}
\end{figure}

\begin{figure}[htbp]
    \centering
    \includegraphics[width=0.5\linewidth]{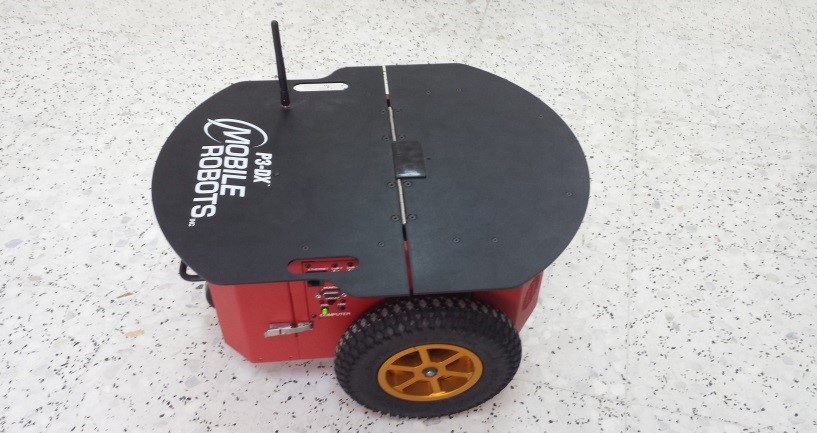}
    \caption{Pioneer P3DX robot}
    \label{fig:robot}
\end{figure}

Figure \ref{fig:experimental_setup} presents the experimental setup utilized in this study. The experiments were carried out using the Pioneer P3DX mobile robot, a well-established and widely adopted platform in robotics research. This robot was employed as the main platform for implementing and testing the proposed path planning approach. A visual of the Pioneer P3DX robot is provided in Figure \ref{fig:robot}.

The path generation step took place on an external workstation equipped with Matlab. The external workstation served as the computational hub, responsible for formulating and optimizing paths before transmitting the generated instructions to the Pioneer P3DX robot. Additionally, the onboard PC of the Pioneer P3DX robot is equipped with Matlab/Simulink software, facilitating the deployment of the generated controller for the robot. The integration of Matlab/Simulink on the onboard PC ensured a seamless execution of the path planning controller, contributing to the overall efficiency of the experimental process.

\begin{figure}[!htbp]
        \begin{center}
            \includegraphics[width=\linewidth]{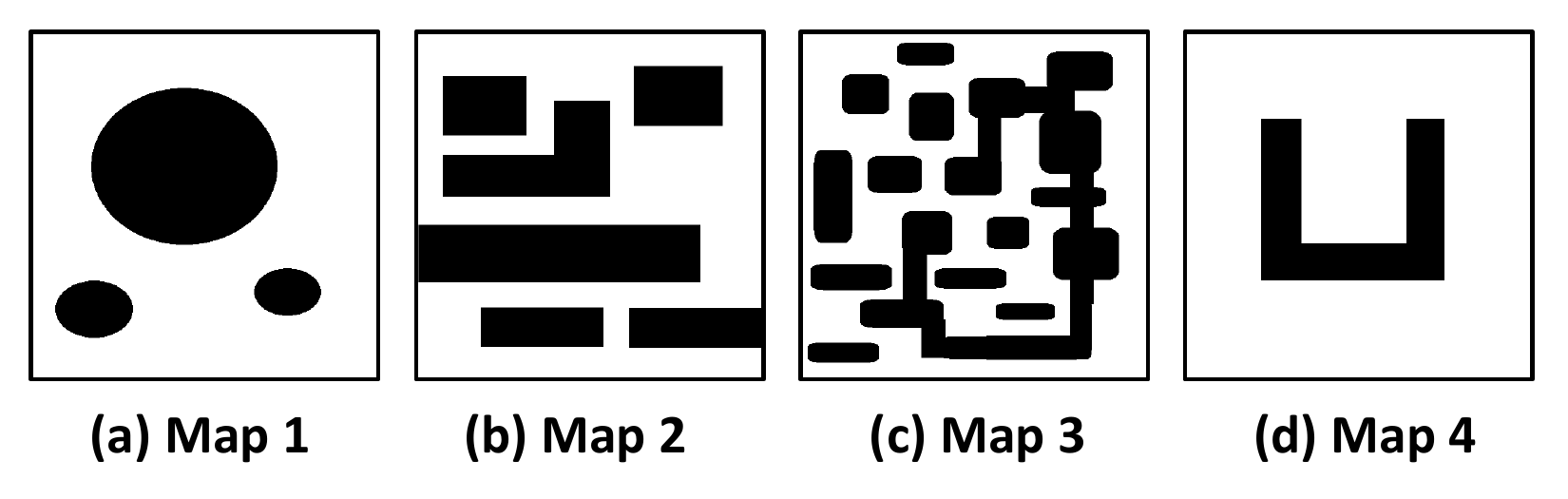}
            \caption{The maps to be used in the experiments}
        	\label{fig:maps}
        \end{center}
    \end{figure}

Figure \ref{fig:maps} illustrates the set of maps used in the experimental evaluation. These maps were carefully selected to assess the performance of the proposed path planning technique under varying levels of complexity and challenge. Map 1 represents a simple environment without the presence of local minima traps. Although straightforward, it serves as a baseline to observe the behavior of the technique in ideal conditions. Map 2 introduces more complex obstacle arrangements that can give rise to potential local minima traps, providing a moderate challenge to test the robustness of the planner. Map 3 features a maze-like structure with numerous local minima and narrow passages, designed to rigorously evaluate the technique’s ability to escape traps and handle constrained navigation spaces. Finally, Map 4 presents a classic U-shaped obstacle configuration, which is widely recognized in path planning literature as a common scenario where standard APF techniques tend to fail due to local minima traps.

To be noted, the unit of path length is expressed in pixels (px). This choice is primarily motivated by the generalization and abstraction of the environment representation. Since all experimental and simulation maps are grid-based and rendered as images, measuring distances in pixels provides a consistent and resolution-independent metric that aligns with the visual representation of the workspace. Using pixels allows for straightforward comparisons between different planning techniques without dependence on real-world scale or units, making it especially useful in benchmarking and algorithm evaluation. However, it is important to note that pixel-based length can be easily scaled to physical units if the resolution and map-to-world scale are known.

\begin{figure}[!htbp]
        \begin{center}
            \includegraphics[width=\linewidth]{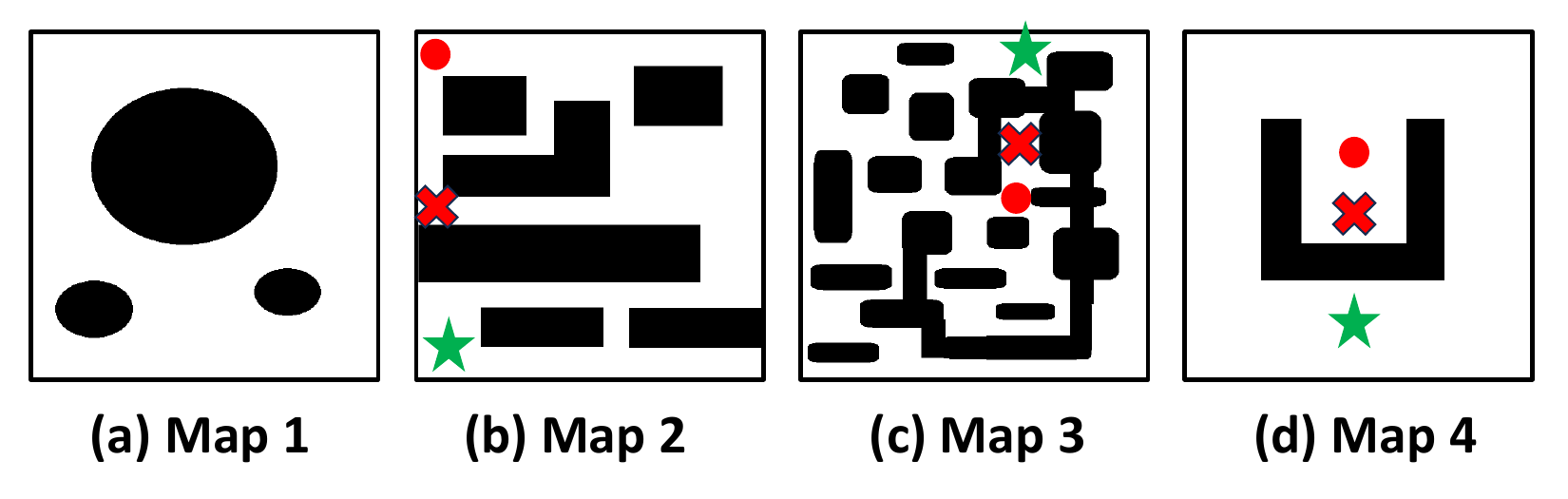}
            \caption{Examples of possible possible local minima traps for some start and goal points in the experimental maps. The green star represents the goal, the red circle indicates the start position, and the red X marks potential local minima traps.}
        	\label{fig:maps_traps}
        \end{center}
    \end{figure}

Figure \ref{fig:maps_traps} illustrates the experimental maps with marked start points, goal points, and potential local minima traps. In each map, the green star symbol indicates the goal position, while the red circle represents the robot’s starting point. The red “X” symbols denote the locations where local minima traps are likely to occur, causing standard APF techniques to fail. In Map 1, there are no local minima, serving as a baseline scenario. In Map 2, a local minima trap is present along the robot’s path to the goal. Map 3 presents a complex maze with multiple local minima traps and narrow passages, significantly challenging the planner. Map 4 highlights the classic U-shaped obstacle configuration where a prominent local minima trap forms in front of the goal area.

\subsection{Escaping local minima}
One of the primary motivations behind the proposed Bulldozer technique is to address the critical limitation of traditional APF methods getting trapped in local minima. First, this section presents a comparative analysis between the standard APF and the proposed method, focusing on their ability to escape such traps and successfully reach the goal. Next, since the proposed technique is specifically designed to address the local minima problem, which is an enhancement that inherently introduces additional computational overhead, it is important to evaluate its impact on processing speed and path length in comparison to competitive APF-based solutions that solve local minima problem.

\subsection{Comparison with standard APF method}
\label{esacpe}

This section compares the performance of the proposed technique compared to standard APF method in terms of the ability to escape local minima. To demonstrate this capability, experiments were conducted using maps specifically designed to induce local minima scenarios, such as Map 2, Map 3, and Map 4. In each case, the robot was initialized in a location prone to entrapment under conventional APF. As expected, the standard APF technique failed to guide the robot toward the goal, resulting in the robot becoming stuck in local minima regions, as illustrated by the marked trap locations in Figure~\ref{fig:maps_traps}.

\begin{figure}[htbp]
    \centering
    \begin{subfigure}[b]{0.3\textwidth}
        \fbox{\includegraphics[width=\textwidth]{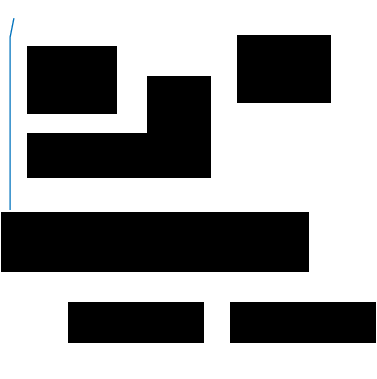}}
        \caption{Map 2 – APF}
    \end{subfigure}
    \hfill
    \begin{subfigure}[b]{0.3\textwidth}
        \fbox{\includegraphics[width=\textwidth]{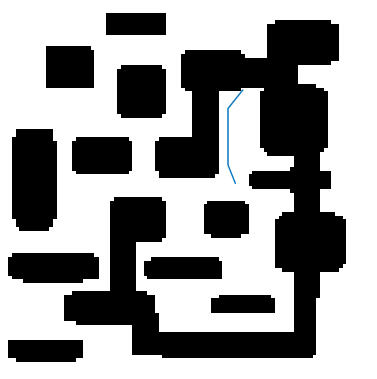}}
        \caption{Map 3 – APF}
    \end{subfigure}
    \hfill
    \begin{subfigure}[b]{0.3\textwidth}
        \fbox{\includegraphics[width=\textwidth]{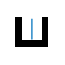}}
        \caption{Map 4 – APF}
    \end{subfigure}
    
    \vspace{0.5cm}
    
    \begin{subfigure}[b]{0.3\textwidth}
        \fbox{\includegraphics[width=\textwidth]{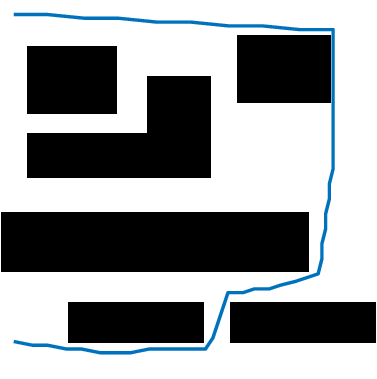}}
        \caption{Map 2 – Proposed}
    \end{subfigure}
    \hfill
    \begin{subfigure}[b]{0.3\textwidth}
        \fbox{\includegraphics[width=\textwidth]{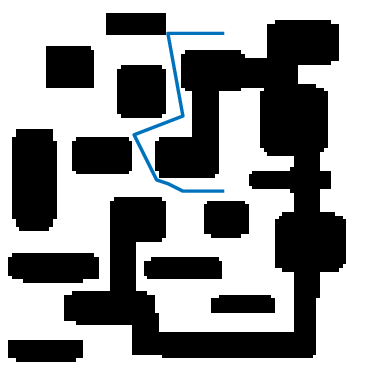}}
        \caption{Map 3 – Proposed}
    \end{subfigure}
    \hfill
    \begin{subfigure}[b]{0.3\textwidth}
        \fbox{\includegraphics[width=\textwidth]{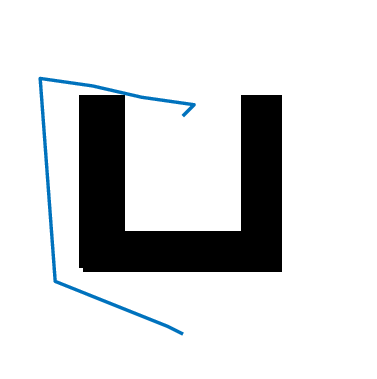}}
        \caption{Map 4 – Proposed}
    \end{subfigure}

    \caption{Experimental results demonstrating the ability of the proposed Bulldozer technique to escape local minima traps. The first row shows the robot's behavior using the standard APF method, where it fails to reach the goal due to entrapment. The second row illustrates the successful navigation of the robot using the proposed technique, which effectively eliminates local minima and enables goal-reaching behavior.}
    \label{fig:escape_comparison}
\end{figure}

In contrast, the Bulldozer technique was able to detect these problematic regions through its iterative backfilling mechanism, which identifies low-potential zones and increases their potential values to eliminate the local minima effect. The robot was subsequently able to continue its motion toward the goal without interruption. The performance comparison between standard APF and the proposed Bulldozer technique is shown in Figure~\ref{fig:escape_comparison}, where it is evident that the proposed method successfully escapes the traps and completes the task in each scenario.

The results clearly show the effectiveness of the proposed technique in overcoming local minima situations where the standard APF approach fails. The Bulldozer method not only ensures trap avoidance but does so without compromising the simplicity and real-time responsiveness that APF is known for as will be proven in next sections.

The superior performance of the proposed technique stems from its ability to actively modify the potential field in response to the presence of local minima. In standard APF, the potential field is static—once generated, it does not adapt to the environment's geometry or the robot's position beyond the initial potential assignments. As a result, when the robot enters a local minimum, the attractive and repulsive forces cancel out, leaving the robot with no guidance to escape the trap.

In contrast, the Bulldozer technique introduces a dynamic mechanism that detects such low-potential regions and progressively increases their potential values. This iterative "backfilling" effectively reshapes the potential field, eliminating the conditions that cause the robot to become stuck. As a result, the robot is continuously guided toward lower potential regions leading to the goal, even in environments where traditional APF fails.

\subsection{Comparison with other solutions}
This section compares the proposed technique with other APF-based local minima solutions published recently in the literature. The first technique will be compared to the proposed technique is an adaptive gain approach proposed in \cite{81}. The second technique to be compared to is based on fluid dynamic concepts \cite{82}.

\subsubsection{A: Comparison with adaptive APF Solution}

This subsection presents a comparative analysis between the proposed Bulldozer technique and an adaptive APF method, as described in \cite{82}, which dynamically adjusts the gains of the repulsive and attractive forces in an attempt to escape local minima traps. While such adaptive strategies can enhance the standard APF performance in certain scenarios, they inherently introduce additional computational overhead due to the continuous search for optimal gain values.

To be noted, the adaptive APF method failed to find a successful path in Maps 2 and 3 when using the original start and goal configurations illustrated in Figure \ref{fig:escape_comparison}. Consequently, modifications to the start and goal locations were necessary to enable the adaptive method to produce a valid path. This already highlights a limitation in robustness compared to the proposed technique, which did not require any changes to the map configuration to successfully complete the task.

\begin{table}[!htbp]
\renewcommand{\arraystretch}{1.3}
\caption{Comparison of path planning techniques: Adaptive APF \cite{82} and the proposed technique in terms of execution speed and the path length}
\label{tbl:comparison_apf}
\centering
\begin{tabular}{l||c|c|c|c} \hline
Map & Technique & Parameter & Length (px) & Time (sec) \\ \hline \hline

\multirow{6}{*}{1} & \multirow{3}{*}{APF \cite{82}} 
& 5\% & 136.75 & 2.24 \\
& & 10\% & 135.25 & 1.08 \\
& & 15\% & 133.75 & 0.74 \\ \cline{2-5}

&  & 5 & 136.14 & 1.41 \\
& Proposed & 10 & 130.22 & 0.81 \\
&  & 15 & 132.78 & 0.62 \\ \hline

\multirow{6}{*}{2} & \multirow{3}{*}{APF \cite{82}} 
& 5\% & 82.75 & 2.36 \\
& & 10\% & 83.25 & 2.09 \\
& & 15\% & 81.50 & 1.86 \\ \cline{2-5}
&  & 5 & 79.97 & 1.02 \\
& Proposed & 10 & 81.44 & 0.87 \\
&  & 15 & 81.90 & 0.82 \\ \hline

\multirow{6}{*}{3} &  & 10\% & 51.75 & 1.95 \\
& APF \cite{82} & 11\% & 51.25 & 1.59 \\
& & 12\% & 51.00 & 1.71 \\ \cline{2-5}

& \multirow{3}{*}{Proposed} & 5 & 59.34 & 1.49 \\
& & 10 & 61.16 & 1.09 \\
& & 15 & 62.68 & 0.96 \\ \hline

\multirow{6}{*}{4} &  & 10\% & 129.75 & 1.94 \\
& APF \cite{82} & 11\% & 128.50 & 1.90 \\
& & 12\% & 129.75 & 1.83 \\ \cline{2-5}

& \multirow{3}{*}{Proposed} & 5 & 113.30 & 1.26 \\
& & 10 & 158.54 & 0.89 \\
& & 15 & 141.55 & 0.80 \\ \hline

\end{tabular}
\end{table}

As shown in Table~\ref{tbl:comparison_apf}, the proposed technique consistently outperformed the adaptive APF in terms of execution speed, despite achieving comparable or even shorter path lengths. For example, in Map 2, the adaptive APF method recorded a best execution time of 1.86 seconds, whereas the proposed method achieved a significantly faster execution time of 0.82 seconds with a similar path length (81.90 px vs. 81.50 px). Similarly, in Map 3, the adaptive APF achieved its best time at 1.59 seconds, while the proposed method recorded a faster time of 0.96 seconds, again with only a slight variation in path length.

This performance gain can be attributed to the operational efficiency of the Bulldozer technique. While adaptive APF spends time iteratively tuning its parameters to escape traps—an inherently time-consuming process—the proposed technique performs a direct modification of the potential field through its backfilling mechanism, eliminating the need for repeated gain adjustment cycles.

Interestingly, while the proposed method was primarily designed to improve robustness and execution speed, it also yielded shorter paths in several cases. For instance, in Map 1, the proposed method achieved a shortest path length of 130.22 px, compared to 135.25 px by the adaptive APF. Similarly, in Map 4, the shortest path generated by the proposed technique was 113.30 px, whereas the adaptive APF paths were consistently longer, exceeding 128 px.

This improvement in path efficiency can be explained by the optimization phase incorporated into the proposed technique, which introduces linear shortcuts into the potential field, smoothing and shortening the resulting path. In contrast, the adaptive APF often produces longer paths due to the influence of stronger repulsive forces used to escape traps, which can push the robot away from the optimal trajectory and lead to more circuitous routes.

\subsubsection{B: Comparison with Fluid Dynamics Solution}

\begin{figure}
    \centering

    \begin{subfigure}[b]{0.45\textwidth}
        \centering
        \fbox{\includegraphics[width=5cm,height=5cm]{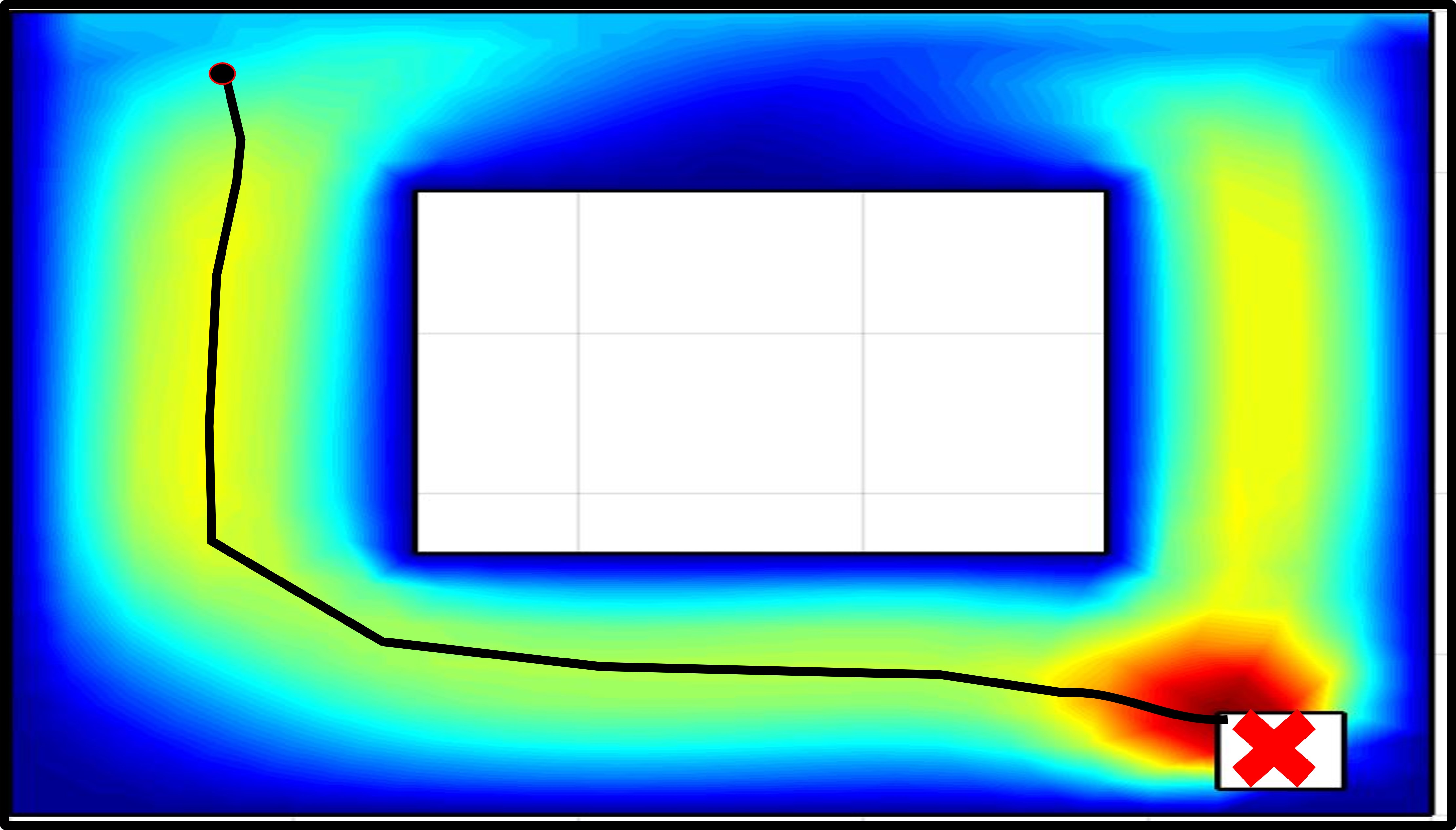}}
        \caption{}
        \label{fig:a}
    \end{subfigure}
    \hfill
    \begin{subfigure}[b]{0.45\textwidth}
        \centering
        \fbox{\includegraphics[width=5cm,height=5cm]{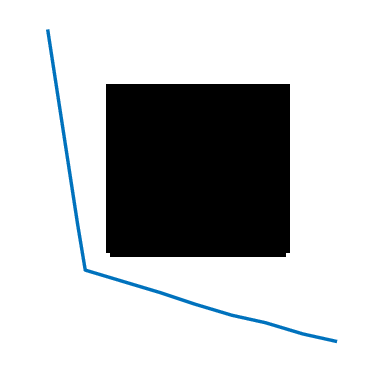}}
        \caption{}
        \label{fig:b}
    \end{subfigure}

    \vspace{0.5cm}

    \begin{subfigure}[b]{0.45\textwidth}
        \centering
        \fbox{\includegraphics[width=5cm,height=5cm]{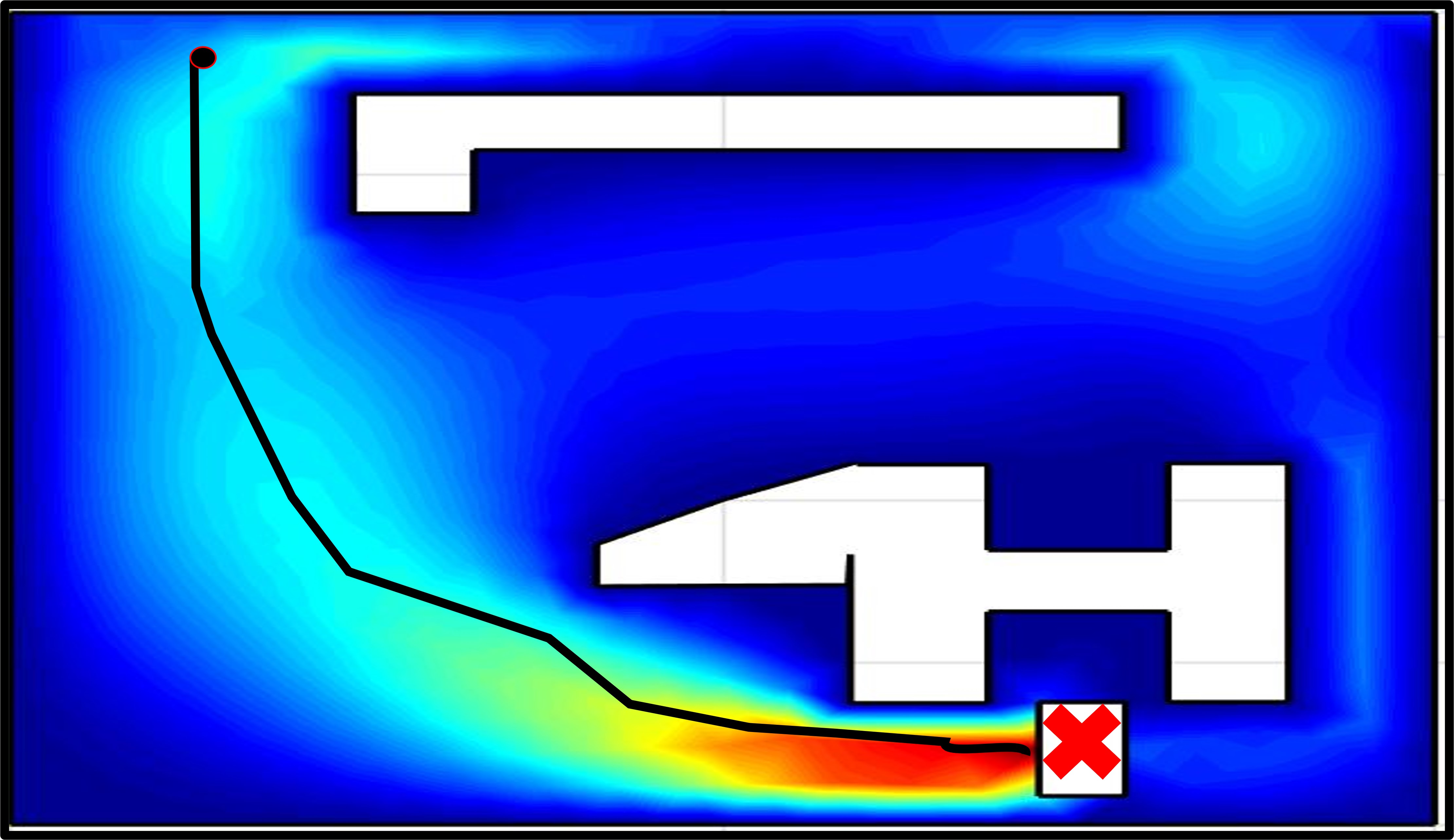}}
        \caption{}
        \label{fig:c}
    \end{subfigure}
    \hfill
    \begin{subfigure}[b]{0.45\textwidth}
        \centering
        \fbox{\includegraphics[width=5cm,height=5cm]{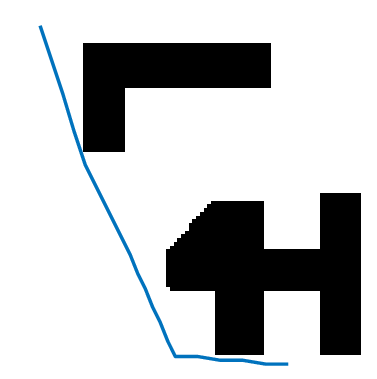}}
        \caption{}
        \label{fig:d}
    \end{subfigure}

    \vspace{0.5cm}

    \begin{subfigure}[b]{0.45\textwidth}
        \centering
        \fbox{\includegraphics[width=5cm,height=5cm]{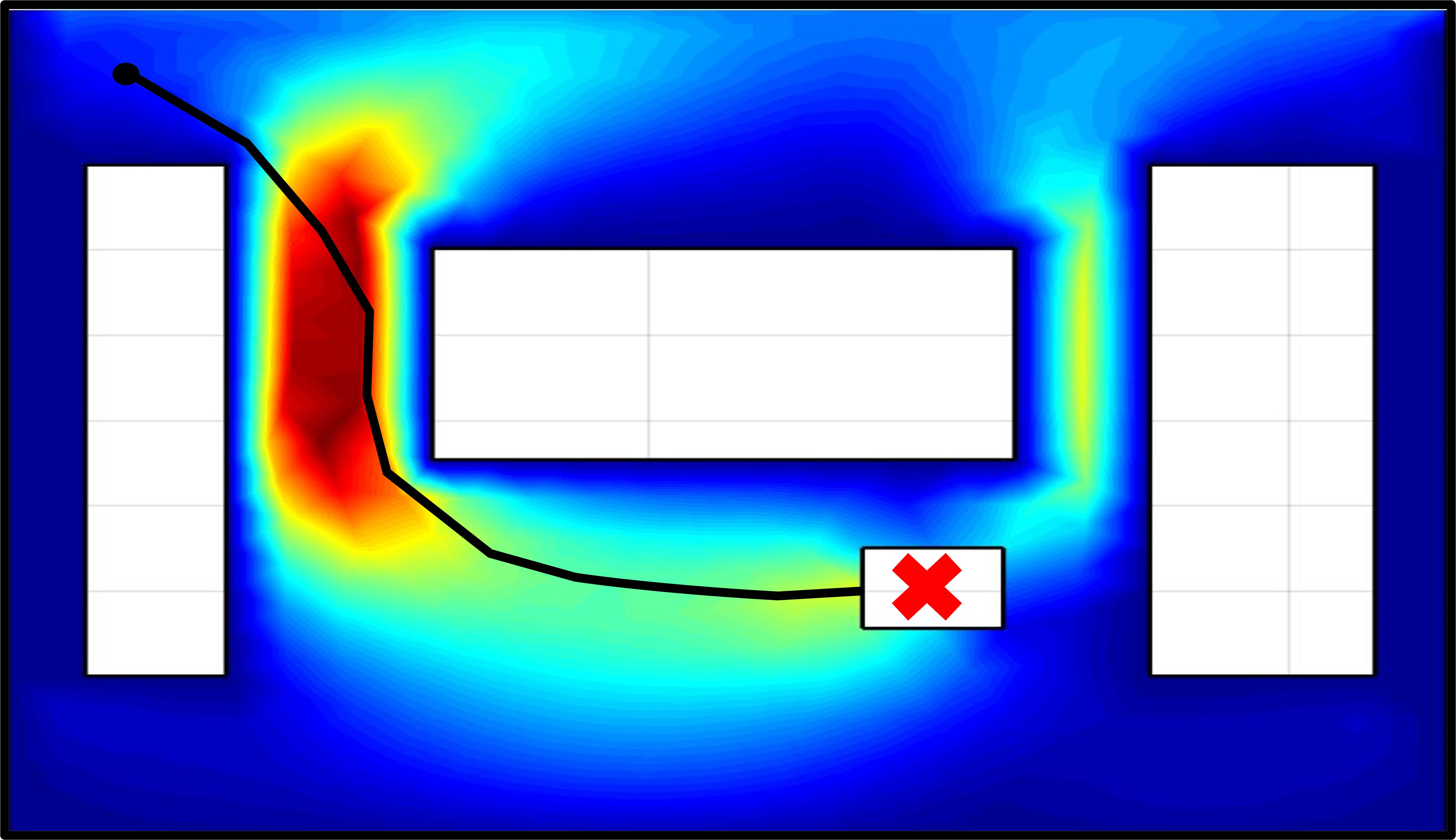}}
        \caption{}
        \label{fig:e}
    \end{subfigure}
    \hfill
    \begin{subfigure}[b]{0.45\textwidth}
        \centering
        \fbox{\includegraphics[width=5cm,height=5cm]{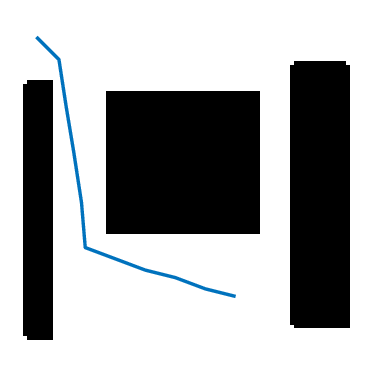}}
        \caption{}
        \label{fig:f}
    \end{subfigure}

    \caption{Showing the paths generated by the fluid dynamics technique \cite{81} in (a), (c), and (e), and the proposed technique (b), (d), and (f). The colors in the fluid dynamics technique refers to the strength of the fluid.}
    \label{fig:fluid}
\end{figure}

This subsection presents a comparative evaluation between the proposed Bulldozer technique and a fluid dynamics-based path planning method \cite{81}. Figure~\ref{fig:fluid} illustrates the generated paths for both techniques across different maps. The left column of Figure~\ref{fig:fluid} shows the trajectories based on fluid dynamics, while the right column presents the paths generated by the proposed technique.

The fluid dynamics-based path planning technique is inspired by the behavior of fluid flow through a navigable space. In this approach, the robot's environment is treated as a fluid domain, and the motion of the robot is guided by the flow of an artificial fluid toward the goal. The technique involves solving the Navier–Stokes equations, which govern the motion of incompressible fluids, to simulate fluid flow within the environment. The robot then follows the streamlines of this flow field, treating the fluid velocity vectors as guidance directions. While this method can effectively avoid obstacles and smooth trajectories, it requires complex modeling and the numerical solution of partial differential equations, which contributes to its computational overhead.

As shown in Table \ref{tbl:fluid_comp}, the proposed technique demonstrates a clear advantage in terms of execution speed. This is primarily due to its computational simplicity, as it builds upon the classical Artificial Potential Field framework with minimal additional processing steps. In contrast, the fluid dynamics method requires the modeling and numerical solution of partial differential equations to simulate fluid flow behavior—an inherently more complex and time-consuming process.

The performance difference is evident in the execution times. For instance, in Map 1, the proposed technique achieved its fastest execution time at 0.72 seconds, whereas the fluid dynamics approach required up to 0.97 seconds. Similarly, in Map 2, the proposed method completed the task in as low as 0.70 seconds, while the fluid-based technique consistently took around 1.10 seconds. These results clearly highlight the lightweight nature of the proposed method, making it more suitable for real-time applications.

In terms of path length, the proposed technique also shows competitive or even superior performance. This advantage is attributed to the direct nature of potential field gradients, which typically point straight toward the goal, resulting in more efficient trajectories. On the other hand, the fluid dynamics technique relies on simulated fluid flow patterns that may be influenced by the geometry of obstacles, potentially creating swirls or tidal effects near obstacle borders. Such phenomena can slightly deflect the path and increase its overall length.

For example, in Map 3, the proposed method achieved a path length of 96.45 px, compared to 107.47 px in the fluid-based approach. Likewise, in Map 1, a minimum path length of 130.22 px was recorded by the proposed technique, which is shorter than the 142.47 px obtained via the fluid dynamics method.

\begin{table} [!htbp]
\renewcommand{\arraystretch}{1.3}
\caption{Comparing the performance of the proposed technique against APF technique in terms of the path length and the execution time}
\label{tbl:fluid_comp}
\centering
\begin{tabular}{l||c|c|c|c} \hline
{Map} & {Technique} & {Parameter} & {Length (px)} & {Time (sec)} \\ \hline

\multirow{6}{*}{1}

& & 50  & 142.47 & 0.92  \\ \cline{3-5}
& Fluid \cite{81} & 100 & 140.84 & 0.97  \\ \cline{3-5}
& & 150 & 141.63 & 0.91  \\ \cline{2-5}

& &5 & 131.12 & 1.29 \\ \cline{3-5}
& Proposed &10 & 140.02 &0.75 \\ \cline{3-5}
& &15 & 134.46 & 0.72 \\ \cline{1-5}

\multirow{6}{*}{2}& & 50  & 112.49 & 1.10  \\ \cline{3-5}
& Fluid \cite{81} & 100 & 112.49 & 1.10  \\ \cline{3-5}
& & 150 & 110.17 & 1.10  \\ \cline{2-5}

& &5 &121.22 &1.28 \\ \cline{3-5}
& Proposed &10 &122.71 &0.93 \\ \cline{3-5}
& &15 &125.47 &0.70 \\ \cline{1-5}

\multirow{6}{*}{3}& & 50  & 107.20 & 1.10  \\ \cline{3-5}
& Fluid \cite{81} & 100 & 104.79 & 1.10  \\ \cline{3-5}
& & 150 & 107.47 & 1.20  \\ \cline{2-5}

& &5 &96.53 &1.51 \\ \cline{3-5}
& Proposed &10 &96.45 &0.86  \\ \cline{3-5}
& &15 &101.12 &0.69 \\ \hline

\end{tabular}
\end{table}

\newpage
\subsection{Comparison with Popular Techniques}
Beyond addressing the limitations of traditional APF methods, it is equally important to evaluate the overall competitiveness of the proposed Bulldozer technique when compared to other well-established and widely adopted path planning algorithms. Popular techniques such as A*, and PRM are commonly used in both academic and industrial applications due to their reliability and robustness in various environments. However, these methods are often associated with higher computational costs and longer execution times, particularly in real-time robotic applications.

Traditionally, APF has been regarded as one of the fastest techniques due to its low computational complexity and continuous guidance mechanism. Nevertheless, its major drawback has always been the local minima trap problem. With the introduction of the Bulldozer technique, which effectively mitigates this limitation, an important question arises: Can APF-based methods now not only retain their speed advantage but also rival or surpass the performance of A*, and PRM in overall path planning effectiveness?

This section explores this question through a series of comparative experiments that assess both execution time and path quality. The results will help determine whether the proposed enhancements to APF allow it to remain the most efficient solution while also improving its robustness and applicability across complex environments.

\subsection{Comparison with Grid-based techniques}

A* (A-star) is a widely used grid-based path planning algorithm that searches for the shortest path between a start and goal point by combining actual travel cost from the start (known as the cost-to-come) with a heuristic estimate of the remaining cost to the goal (cost-to-go). It guarantees optimality in path planning on a discretized map and is known for its balance between efficiency and accuracy.

Since A* operates on a discretized grid, its performance is highly dependent on the resolution of the grid. A finer resolution (meaning smaller grid cells) enables the planner to produce more accurate and smoother paths. However, this comes at the cost of increased computational time, as the number of nodes to search increases significantly. On the other hand, a coarser grid leads to faster computation but at the expense of path quality and potentially suboptimal routes. Therefore, a clear trade-off exists between path optimality and execution speed when choosing the resolution parameter for A*.

The results in Table~\ref{tbl:comparison_a} compare the proposed Bulldozer technique with A* across various map scenarios using different resolution settings. In Map 1, A* achieved a shortest path length of 135.50 px at a resolution of 250, but the execution time increased substantially to 2.468 seconds. In contrast, the proposed technique achieved a slightly shorter path of 130.22 px with a much faster execution time of 0.81 seconds, demonstrating the computational efficiency advantage of the proposed method.

\begin{table}[!htbp]
\renewcommand{\arraystretch}{1.3}
\caption{Comparison of path planning techniques: A* and the proposed technique in terms of execution speed and the path length (all lengths scaled to 100×100 map)}
\label{tbl:comparison_a}
\centering
\begin{tabular}{l||c|c|c|c} \hline
Map & Technique & Parameter & Length (px) & Time (sec) \\ \hline \hline

\multirow{6}{*}{1} &  & 150 & 136.00 & 0.816 \\
& A* & 200 & 135.75 & 1.353 \\
&  & 250 & 135.50 & 2.468 \\ \cline{2-5}
&  & 5 & 136.14 & 1.41 \\
& Proposed & 10 & 130.22 & 0.81 \\
&  & 15 & 132.78 & 0.62 \\ \hline


\multirow{6}{*}{2} & & 200  & 131.24 & 0.93\\
  & A* & 250 & 131.07 & 1.66 \\
  & & 300 & 130.76 & 2.72 \\ \cline{2-5}
  
&  & 5 & 123.12 & 1.06 \\
& Proposed & 10 & 125.78 & 0.885 \\
& & 15 & 125.71 & 0.85 \\ \hline

\multirow{6}{*}{3} & & 200 & 109.00 & 0.73 \\
  & A* & 250 & 108.64 & 1.02 \\
  & & 300 & 107.54 & 1.57 \\ \cline{2-5}
  
& \multirow{3}{*}{} & 5 & 102.98 & 1.56 \\
  & Proposed& 10 & 103.75 & 0.97 \\
  & & 15 & 106.43 & 0.93 \\ \hline

\multirow{6}{*}{4} & & 150 & 105.75 & 0.89 \\
  & A* & 200 & 103.75 & 1.21 \\
  & & 250 & 103.75 & 1.94 \\ \cline{2-5}
& \multirow{3}{*}{} & 5 & 113.30 & 1.26 \\
  & Proposed& 10 & 158.54 & 0.89 \\
  & & 15 & 141.55 & 0.80 \\ \hline  
\end{tabular}
\end{table}

Similarly, in Map 2, A* produced a path of 130.76 px at its highest resolution, requiring 2.72 seconds, while the proposed technique generated a comparable path of 125.71 px in just 0.85 seconds. The performance gap in terms of speed becomes more evident as the A* resolution increases.

In Map 3, the proposed technique achieved a path of 103.75 px in 0.97 seconds, whereas A* required 1.57 seconds to generate a similar path of 107.54 px. These results further highlight that the proposed method can compete with A* in terms of path quality while consistently offering superior execution times.

Interestingly, even in Map 4, the proposed method managed to complete the path faster in all cases, with a minimum time of 0.80 seconds, while A* took up to 1.94 seconds depending on the resolution. The path lengths remained close in value, demonstrating that the Bulldozer-enhanced APF retains good path quality even in complex environments.

The significant speed advantage of the proposed technique can be attributed to its nature—being fundamentally an enhanced APF method. It benefits from real-time responsiveness without the overhead of node expansion, open list management, and grid traversal operations inherent to A*. Moreover, the proposed technique incorporates an optimization phase that smooths and shortens the path using linear shortcuts, which helps achieve efficient trajectories without increasing computational complexity.

\subsection{Comparison with Sampling-based}

This subsection presents a comprehensive comparison between the proposed Bulldozer technique and the PRM method. PRM is a sampling-based path planning technique that builds a roadmap of feasible paths in the configuration space by randomly sampling collision-free points (nodes) and connecting them through valid paths (edges). Once the roadmap is constructed, the start and goal positions are connected to the nearest nodes, and a graph search algorithm is used to find a path.

One of the most critical factors influencing PRM performance is the number of sampled nodes, which directly affects the connectivity and quality of the roadmap. A higher number of samples typically improves the likelihood of finding shorter and smoother paths due to better coverage of the configuration space. However, this improvement comes at a significant computational cost. More samples lead to an increase in the number of edges and connections that must be evaluated and stored, resulting in longer execution times during both the roadmap construction and the subsequent path search.

The results shown in Table~\ref{tbl:comparison_prm} clearly reflect this trade-off. For example, in Map 1, increasing the number of PRM samples from 150 to 250 resulted in a modest improvement in path length—from 133.45 px to 133.50 px—but a substantial rise in execution time—from 3.59 seconds to 8.25 seconds. This highlights the diminishing return effect of adding more samples: while the path quality may plateau, the computational burden continues to rise significantly.

\begin{table}[!htbp]
\renewcommand{\arraystretch}{1.3}
\caption{Comparison of path planning techniques: PRM and the proposed technique in terms of execution speed and path length}
\label{tbl:comparison_prm}
\centering
\begin{tabular}{l||c|c|c|c} \hline
Map & Technique & Parameter & Length (px) & Time (sec) \\ \hline \hline

\multirow{6}{*}{1} 
& & 150 & 133.45 & 3.59 \\
& PRM & 200 & 133.13 & 5.97 \\
& & 250 & 133.50 & 8.25 \\ \cline{2-5}
& & 5 & 136.14 & 1.41 \\
& Proposed & 10 & 130.22 & 0.81 \\
& & 15 & 132.78 & 0.62 \\ \hline

\multirow{6}{*}{2} 
& & 150 & 128.63 & 2.61 \\
& PRM & 200 & 126.15 & 3.54 \\
& & 250 & 124.78 & 5.18 \\ \cline{2-5}

& & 5 & 123.12 & 1.06 \\
& Proposed & 10 & 125.78 & 0.89 \\
& & 15 & 125.71 & 0.85 \\ \hline

\multirow{6}{*}{3} 
& & 150 & 109.00 & 1.05 \\
& PRM & 200 & 109.00 & 1.74 \\
& & 250 & 109.00 & 2.36 \\ \cline{2-5}

&  & 5 & 102.98 & 1.56 \\
& Proposed& 10 & 103.75 & 0.97 \\
& & 15 & 106.43 & 0.93 \\ \hline

\multirow{6}{*}{4} 
& & 150 & 117.68 & 2.36 \\
& PRM & 200 & 117.42 & 3.83 \\
& & 250 & 116.93 & 6.26 \\ \cline{2-5}

&  & 5 & 113.30 & 1.26 \\
& Proposed & 10 & 158.54 & 0.89 \\
& & 15 & 141.55 & 0.80 \\ \hline

\end{tabular}
\end{table}

In contrast, the proposed Bulldozer technique delivers consistently better performance in both path length and processing speed. For instance, in Map 1, the proposed method achieved a shorter path length of 130.22 px with an execution time of only 0.81 seconds, far outperforming PRM at all sampling levels. Similarly, in Map 2, the proposed technique produced a path length of 125.71 px at 0.85 seconds, while PRM with 250 samples achieved a similar path length of 124.78 px, but at the cost of 5.18 seconds of computation time.

These differences become even more pronounced in Map 3, where PRM generated the same path length of 109.00 px across all sampling levels, despite increased computation time from 1.05 seconds to 2.36 seconds. This suggests that for certain environments with constrained connectivity, PRM's roadmap does not significantly benefit from additional samples. Meanwhile, the proposed Bulldozer method demonstrated path lengths of 102.98 px to 106.43 px with faster execution times ranging from 0.93 to 1.56 seconds, confirming its efficiency and adaptability even in complex scenarios.

In Map 4, which features a challenging U-shaped obstacle, PRM required 6.26 seconds to achieve a path length of 116.93 px using 250 samples, whereas the proposed technique achieved a slightly longer but still competitive path length of 141.55 px in just 0.80 seconds. This further underscores the computational efficiency of the proposed approach, especially when real-time performance is prioritized.

Overall, the analysis shows that while PRM is a powerful planner in terms of robustness and global search capabilities, it suffers from increased processing times and diminishing improvements with excessive sampling. The proposed Bulldozer technique, on the other hand, achieves a highly favorable balance between speed and path quality by leveraging the simplicity of APF, enhanced with local minima handling and path optimization features. The absence of a heavy sampling and connection phase makes it inherently faster and more suitable for time-critical robotic applications. Thus, the proposed method proves to be not only an enhancement over traditional APF but also a strong competitor to advanced sampling-based planners such as PRM.

\subsection{Comparison with Other Meta-Heuristic Optimization Techniques}

To evaluate the suitability of different meta-heuristic optimization methods for identifying the lowest potential values during the planning phase, a comparative analysis was conducted using PSO, GA, ABC, and GWO. These techniques were evaluated based on two performance metrics: the final path distance and the execution time required to converge to the optimal local decision at each step.

\begin{figure}[h]
    \centering
    \begin{subfigure}{0.48\textwidth}
        \centering
        \includegraphics[width=\linewidth]{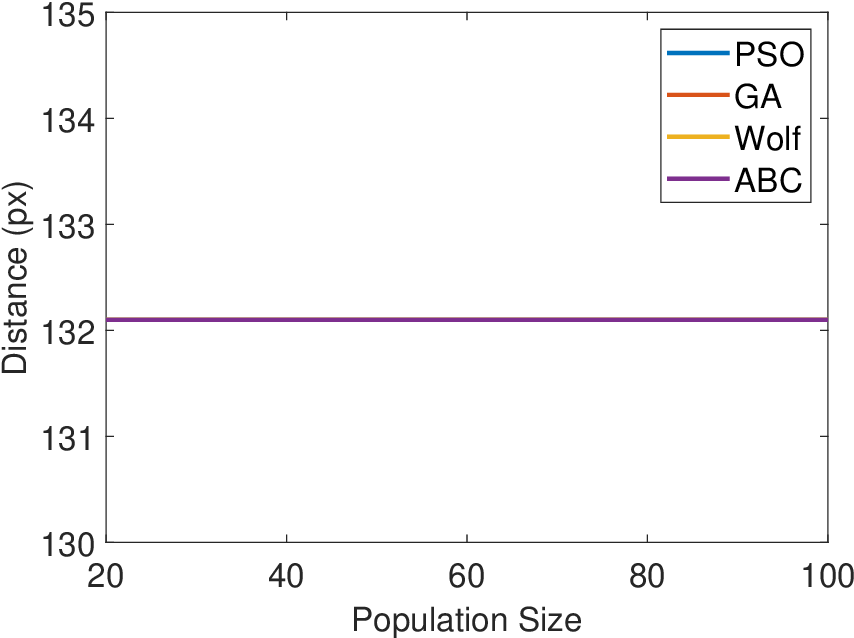}
        \subcaption*{(a)}
        \label{fig:meta_distance}
    \end{subfigure}
    \hfill
    \begin{subfigure}{0.48\textwidth}
        \centering
        \includegraphics[width=\linewidth]{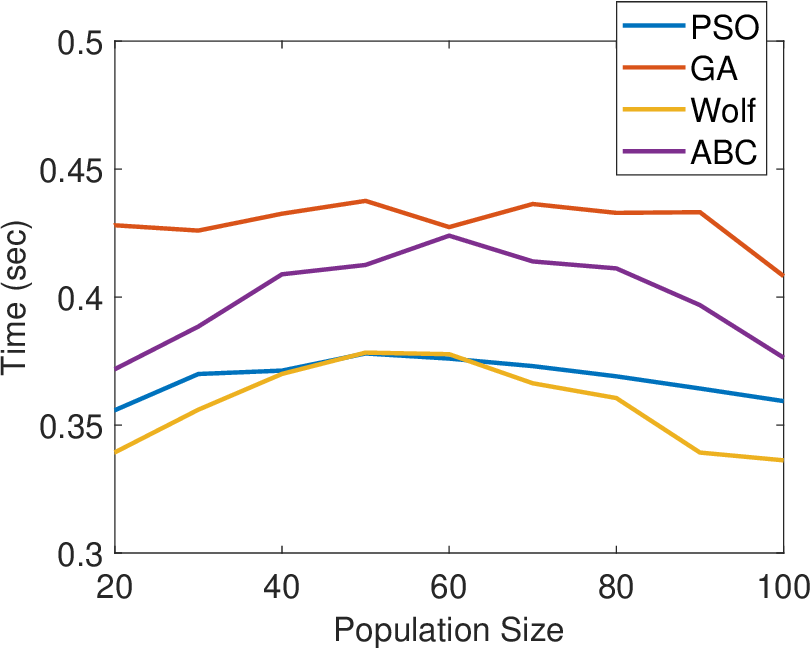}
        \subcaption*{(b)}
        \label{fig:meta_time}
    \end{subfigure}
    \caption{Performance comparison of meta-heuristic optimization techniques under varying population sizes: (a) resulting path distance and (b) execution time.}
    \label{fig:meta_comparison}
\end{figure}

Figure~\ref{fig:meta_comparison}-a shows the distance values obtained by each method across different population sizes ranging from 10 to 100. It can be observed that all techniques produced nearly identical path lengths, indicating that the optimization process was consistently able to find the global or near-optimal solution. This is largely attributed to the fact that the search space in this specific application is relatively small, as it is limited to the robot’s surrounding blocks. Therefore, the population parameter does not significantly influence the path length.

However, the execution time results, shown in Figure~\ref{fig:meta_comparison}-b, reveal substantial differences between the methods. Among all techniques, GWO achieved the best performance in terms of computation time. This can be attributed to GWO's fast convergence behavior and efficient balance between exploration and exploitation, especially in constrained local search spaces.

On the other hand, the GA technique exhibited the highest computation time. This is likely due to its reliance on multiple stages such as selection, crossover, and mutation, which introduce additional processing overhead, especially as the population size increases. ABC was the second slowest, as its iterative foraging behavior involves repeated evaluations and probabilistic decisions that are not always optimal early in the search. PSO performed comparably to GWO, maintaining a balance between simplicity and effectiveness in terms of speed.

Another interesting observation from Figure~\ref{fig:meta_comparison}-b is that all techniques exhibited a noticeable increase in execution time for mid-range population sizes (approximately between 30 and 70). This suggests a trade-off behavior: smaller populations result in fewer computations, while larger populations, despite involving more particles or agents, often converge faster due to the richer diversity. Mid-range population sizes, however, lack both benefits—being neither small enough to compute quickly nor large enough to ensure fast convergence.

While all methods were successful in guiding the robot toward the goal with minimal path variation, GWO demonstrated the best computational efficiency, making it the most suitable choice for real-time embedded implementations of the proposed path planning technique.

\subsection{Traceability: Smoothness Attribute}

In this section, we examine the traceability of the paths generated by the proposed Bulldozer technique using a simple kinematic tracking controller. The objective is to evaluate whether the generated trajectories are smooth and feasible enough to be tracked by a real robot without introducing abrupt control inputs or oscillations.

The tracking was performed using a well-known kinematic controller based on Kanayama’s transformation. The control input vector \( V(t) \), which consists of the linear and angular velocities, is defined as:

\begin{equation}
V(t)=\begin{bmatrix}
v\\
\omega
\end{bmatrix} = \begin{bmatrix}
k_1\tilde{x}+v_d\cos\tilde{\phi}\\
w_d+k_2v_d\tilde{y}+k_3v_d\sin\tilde{\phi}
\end{bmatrix}
\end{equation}
where \( \tilde{x} \), \( \tilde{y} \), and \( \tilde{\phi} \) represent the tracking errors calculated using the Kanayama transformation. The terms \( v_d \) and \( w_d \) denote the desired forward and angular velocities, respectively. The parameters \( k_1 > 0 \), \( k_2 > 0 \), and \( k_3 > 0 \) are the controller gains that govern the convergence behavior and response smoothness. Further derivations, stability proofs, and experimental validation of this controller are provided in \cite{fareh2019vision}.

\begin{figure}[htbp]
    \centering
    \includegraphics[width=\textwidth]{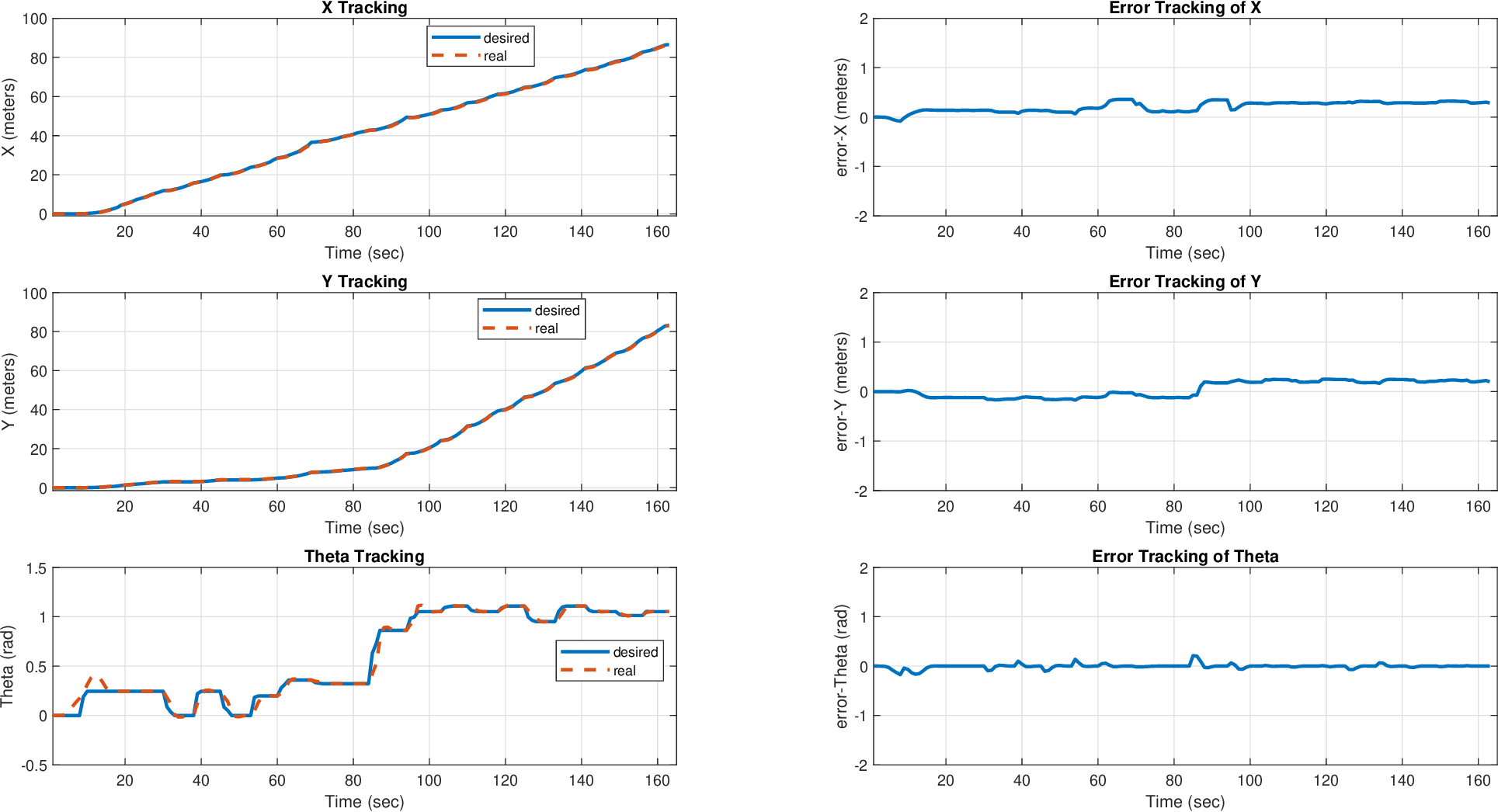}
    \caption{Tracking performance of the proposed path using a simple kinematic controller: (a) X Tracking, (b) X Error, (c) Y Tracking, (d) Y Error, (e) Orientation (\( \theta \)) Tracking, (f) Orientation Error.}
    \label{fig:smoothness_tracking}
\end{figure}

Figure~\ref{fig:smoothness_tracking} illustrates the tracking performance of the proposed paths. Subfigures (a), (c), and (e) depict the actual tracking of X, Y, and orientation (\( \theta \)) compared to the desired trajectory, while subfigures (b), (d), and (f) show the corresponding tracking errors.

It is clearly observed that the robot successfully tracks the reference trajectory with minimal deviation and without significant oscillations. The maximum error in X and Y remains well within acceptable bounds (less than 0.3 meters), and the heading angle error is smoothly corrected over time, demonstrating the controller’s effectiveness.

The smooth tracking performance can be attributed to two key aspects of the proposed Bulldozer technique. First, the potential field inherently produces continuous and smooth gradients, unlike grid-based or sampling-based methods that often result in abrupt, blocky paths. Second, the path optimization phase embedded in the Bulldozer method further refines the trajectory by applying linear shortcuts that reduce unnecessary waypoints and smoothen the path curvature. As a result, the final trajectory becomes more natural and kinematically feasible for the robot to follow.

These results confirm that the proposed technique not only offers efficient path planning but also ensures high-quality trajectories that are well-suited for real-world robotic execution without the need for additional smoothing or replanning stages.

\subsubsection{Quantitative Tracking Performance}

Table~\ref{tab:tracking_performance} presents the quantitative tracking performance of the distributed control scheme in joint space. The tracking accuracy is evaluated using the maximum absolute error, mean absolute error (MAE), root mean square error (RMSE), and final steady-state error for the $X$, $Y$, and $\Theta$ states.

\begin{table}[h]
\centering
\caption{Quantitative tracking performance of the distributed control scheme.}
\label{tab:tracking_performance}
\begin{tabular}{lcccc}
\hline
State & Max Abs. Error & MAE & RMSE & Final Error \\
\hline
$X$     & 0.3597 & 0.2104 & 0.2324 & 0.2867 \\
$Y$     & 0.2496 & 0.1523 & 0.1679 & 0.1992 \\
$\Theta$ & 0.2111 & 0.0218 & 0.0464 & $1.42 \times 10^{-4}$ \\
\hline
\end{tabular}
\end{table}

For the $X$ position, the controller achieves a maximum absolute error of $0.3597~\mathrm{m}$, with a mean absolute error of $0.2104~\mathrm{m}$ and an RMSE of $0.2324~\mathrm{m}$. The final steady-state error converges to $0.2867~\mathrm{m}$, indicating stable tracking behavior with bounded transient deviations.

In the $Y$ direction, improved tracking performance is observed. The maximum absolute error is reduced to $0.2496~\mathrm{m}$, while the MAE and RMSE reach $0.1523~\mathrm{m}$ and $0.1679~\mathrm{m}$, respectively. The final error settles at $0.1992~\mathrm{m}$, confirming consistent convergence.

For the orientation $\Theta$, the controller demonstrates significantly higher accuracy. The maximum absolute error is limited to $0.2111~\mathrm{rad}$, with a low MAE of $0.0218~\mathrm{rad}$ and an RMSE of $0.0464~\mathrm{rad}$. Notably, the final steady-state error is nearly zero ($1.42 \times 10^{-4}~\mathrm{rad}$), indicating excellent asymptotic convergence in angular tracking.

Overall, these results confirm that the proposed distributed control strategy achieves reliable trajectory tracking with low steady-state errors, particularly in the orientation component, while maintaining acceptable tracking accuracy in translational motion.

\section{Conclusions}\label{section-9-conclusions}

This paper presented a novel path planning technique, termed the \textit{Bulldozer}, aimed at overcoming the local minima trap problem associated with traditional APF methods. While APF techniques are widely recognized for their simplicity, low computational cost, and real-time applicability, their susceptibility to local minima has limited their use in complex environments. The proposed Bulldozer method preserves the benefits of APF while introducing a systematic approach to eliminate local minima. This is achieved through a block-wise segmentation of the potential field, followed by an iterative backfilling mechanism that raises the potential values of detected local minima areas. A unique enhancement, the \textit{ramp} technique, was introduced to assist robots that begin their motion from within a trap, allowing a gradual potential descent out of the local minima region. The system was experimentally validated on a physical robot across a variety of environments, ranging from simple to highly cluttered maps. Comparative studies demonstrated the superiority of the proposed technique in terms of robustness and execution time when compared to standard APF and adaptive APF approaches. Furthermore, it outperformed popular path planning algorithms such as A* and PRM in computational speed while maintaining competitive path quality. A kinematic tracking controller was used to evaluate the traceability and smoothness of the generated paths, with results confirming their real-world feasibility. Additionally, the system incorporated GWO algorithm to refine path points within each selected block, leveraging GWO’s fast convergence to improve the efficiency of the planning phase. This highlighted the method’s modularity and ability to integrate optimization algorithms effectively. As this study is
implemented in static environment, in the future the method will be
improved to handle planning paths in dynamic environments.


\bibliographystyle{apa}
\bibliography{ref.bib}

\end{document}